\documentclass{article}
\usepackage[preprint]{corl_2026} 

\usepackage{booktabs,tabularx}
\usepackage{graphicx}
\usepackage{subcaption}
\usepackage{multirow}
\usepackage{amsmath}
\usepackage{amssymb} 
\usepackage{wrapfig}
\usepackage{tablefootnote}
\usepackage[dvipsnames]{xcolor}
\usepackage{enumitem}
\usepackage[symbol]{footmisc}
\usepackage{threeparttable}
\usepackage{makecell}
\usepackage{tikz}
\usetikzlibrary{calc}
\usepackage[normalem]{ulem} 
\usepackage{bm}
\usepackage{siunitx}
\usepackage{booktabs}
\usepackage[toc,page,header]{appendix}
\usepackage{minitoc}
\usepackage{cprotect}
\usepackage[table]{xcolor}

\setlist{nosep} 

\usepackage[toc,page,header]{appendix}
\usepackage{minitoc}

\definecolor{cornflowerblue}{RGB}{100, 149, 237}
\definecolor{orange}{RGB}{230, 159, 0}
\definecolor{lightseagreen}{RGB}{32, 178, 170}
\definecolor{tostEqBg}{HTML}{C8E6C9}
\definecolor{tostEqFg}{HTML}{1B5E20}
\definecolor{tostDiffBg}{HTML}{FFCDD2}
\definecolor{tostDiffFg}{HTML}{B71C1C}
\definecolor{tostIncBg}{HTML}{FFE0B2}
\definecolor{tostIncFg}{HTML}{7A4F01}
\definecolor{onestep}{HTML}{d1495b}
\definecolor{nstep}{HTML}{1f6f8b}

\title{When One Step Is Enough: An Empirical Study of Euler Integration in Flow-Matching Policies}

\author{
Shivakanth Sujit$^{*1}$, Luca Nunziante$^{*1}$, Rousslan Fernand Julien Dossa$^{1}$, Kai Arulkumaran$^{1}$ \\
${}^{1}$Araya Inc., Tokyo\\
\texttt{\small <sujit$\_$shivakanth, luca$\_$nunziante, dossa, kai$\_$arulkumaran>@araya.org}\\
\href{https://onestepflow.github.io}{onestepflow.github.io}
}

\begin{document}
\doparttoc 
\faketableofcontents 
\def\thefootnote{*}\footnotetext{Equal contribution}\def\thefootnote{\arabic{footnote}}
\def\thefootnote{}\footnotetext{Correspondence to \texttt{\small luca$\_$nunziante@araya.org}
}\def\thefootnote{\arabic{footnote}}

\maketitle
\begin{abstract}
Flow matching is now the dominant approach for vision-language action (VLA) robot policies, thanks to its expressiveness and simple training objective.
However, inference for standard flow matching requires iteratively solving an ODE, limiting the control frequency at deployment.
While several works have proposed novel methods to allow for single-shot action generation, we ask whether such additional objectives are actually necessary.
We find that across six flow matching VLAs, one-step action generation largely matches multi-step sampling at scale: multi-step integration confers little to no consistent advantage at inference. We empirically find that this holds as long as models are sufficiently trained and have enough capacity.
Practically, the implication is that for large VLAs, the single-step policy comes essentially for free from the multi-step model, enabling low-latency deployment without retraining, distillation, or auxiliary objectives.
\end{abstract}

\section{Introduction}

\begin{figure}[htp]
\centering
    \begin{subfigure}{\textwidth}
        \centering
        \includegraphics[width=\linewidth]{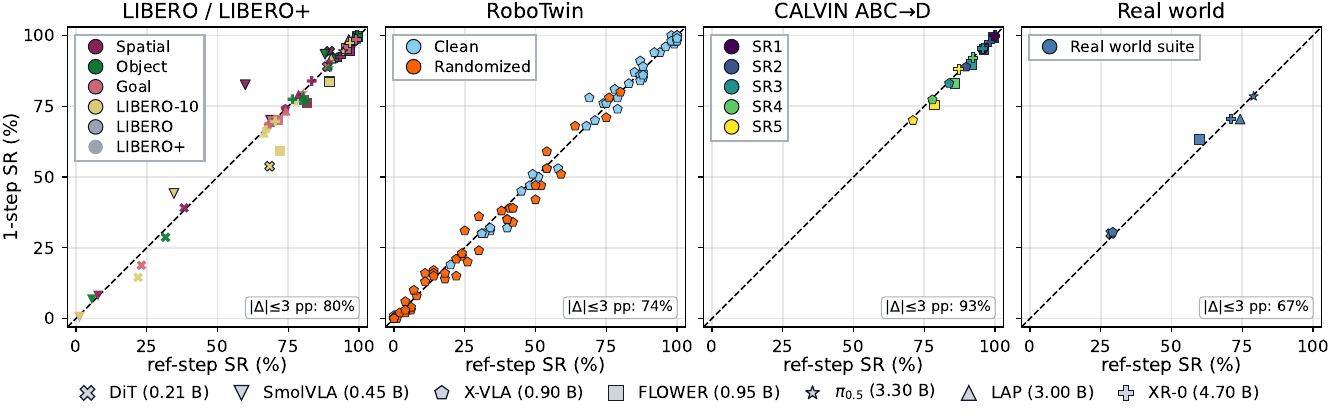}
    \end{subfigure}
    \caption{1-step vs. reference-step success rates (SR) across all (model, benchmark) cells in our study; $\Delta=\text{SR}_\text{ref}-\text{SR}_{1}$ is the performance gap in percentage points (pp).
    Points on or near the x=y diagonal indicate a negligible performance gap between the two conditions.
    For the three larger models ($\pi_{0.5}$, LAP, XR-0) all reported gaps are small and either statistically equivalent under two one-sided t-tests (TOST) or inconclusive with one-step having a higher aggregate SR.
    In the real world pooled deltas $\Delta$ are at most $3.7\,$pp, although with some task-level differences.}
    \label{fig:fig1-panels}
\end{figure}

Diffusion and flow matching models have now become the common choice for training robotic control policies for manipulation~\cite{chi2025diffusion, wang2024rise,zhang2025flowpolicy,park2025flow,jones2025multitaskditpolicy}, consistently outperforming simple regression/Gaussian baselines. To generate action samples, these models learn a vector field that drives noise samples, taken from a source distribution, to a target distribution. Of these, flow matching is now the dominant paradigm by virtue of its simpler form (solving an ordinary differential equation (ODE) instead of a stochastic differential equation (SDE)). Many of the latest robot foundation models, in the form of vision-language-action (VLA) models, combine a large, pretrained vision-language model (VLM) backbone~\cite{marafioti2025smolvlm,xiao2024florence,beyer2024paligemma,bai2025qwen3}, providing existing image and text understanding, with a flow matching action head~\cite{shukor2025smolvla,zheng2025xvla,reuss2025flower,black2025pi05,zha2026lap,cai2026xiaomi}.

In flow matching policies, a velocity field is learned and generating actions requires integrating the learned velocity field, typically with an Euler integrator that evaluates the network $N$ times. 
Higher $N$ results in lower discretization error at the cost of increased latency, thus limiting control frequency, and subsequently hurting fluid motion in dynamic or contact-rich tasks. 
Recent works mitigate these issues by distilling the multi-step policy into a single-step proxy~\cite{park2025flow}, introducing distribution-shaping objectives that are expensive for large-scale VLAs~\cite{li2026q0learning}, or augmenting the flow matching objective to encourage straight-line flows that need fewer integration steps~\cite{prasad2024consistency,yan2025maniflow0,frans2025one,geng2026mean}.
This multi-step formulation also necessitates backpropagation-through-time (BPTT) for gradient-based optimization of reinforcement learning (RL) objectives, which is costly and prone to unstable gradients.

Due to the simplicity of its formulation, in practice VLAs are trained with the \emph{rectified} flow matching objective)~\cite{liu2022flow}, where the target velocity along each training pair is constant, producing linear interpolation paths between source and target samples. It is not the case in general that the learned trajectories are straight, as intersecting linear paths produce curved learned trajectories, which in image generation degrades single-step sample quality~\cite{liu2022flow}. 
However, for robot control, given the current observation and state, the nature of the conditional target distribution may differ substantially from the image generation setting. 
Prior work~\cite{pan2026much} finds that, for control policies, the performance advantage of generative models over regression ones mainly comes from stochasticity injection and the dense supervision signal during training, rather than from modeling multimodality, which they find largely absent in robotics datasets (including the ones understood to be multimodal like \verb|Push T|), likely due to their high-dimensional observation space. 
Concurrent work~\cite{chen2026let} reaches a similar conclusion for VLAs, but studies a single architecture, focusing on training hyperparameters on LIBERO~\cite{liu2023libero} and its variants, with a small real-world experiment. In contrast, we study six VLAs across multiple benchmarks, run a large-scale real-world evaluation, and analyse the smoothness and diversity of the generated trajectories.

We ask whether large-scale VLAs trained with the rectified flow objective and no additional regularization can be integrated in a single step at inference without loss of performance. 
We answer this question with what is, to our knowledge, the first paired one-step vs multi-step ablation of flow matching VLAs, covering six VLAs across multiple benchmarks. 
We evaluate six flow matching VLA models spanning $450$M to $4.7$B parameters (SmolVLA~\cite{shukor2025smolvla}, FLOWER~\cite{reuss2025flower}, X-VLA~\cite{zheng2025xvla}, $\pi_{0.5}$~\cite{black2025pi05}, LAP~\cite{zha2026lap}, XR-0~\cite{cai2026xiaomi}) on four simulation benchmarks (LIBERO~\cite{liu2023libero}, LIBERO+~\cite{fei2025liberop}, CALVIN~\cite{mees2022calvin}, RoboTwin 2.0~\cite{chen2025robotwin}) and a real world Franka suite~\cite{zhang2026augmented}.
All comparisons use matched seeds and real world paired testing. We compare success rates through the two one-sided t-tests (TOST) equivalence test~\cite{schuirmann1987comparison}, analyse the difference in smoothness and diversity of the generated trajectories and also report a latency comparison. Results are summarised in Fig.~\ref{fig:fig1-panels}. Our findings are as follows:

\begin{itemize}
    \item \textbf{Step-equivalence is a factor of model scale, VLM pretraining and training maturity:} The four largest VLAs show strong robustness to single-step inference across simulation benchmarks. $\pi_{0.5}$ and XR-0 are statistically equivalent within $3$ percentage points (pp) on every simulation benchmark for which a public checkpoint was available. X-VLA and LAP are equivalent on all but one benchmark each (Sec.~\ref{subsec:sr-analysis}).
    Initialization ablations further suggest that pretrained visual-language representations favour step-equivalence, even without robot-action pretraining, and the difference between one-step and multi-step inference fades with longer fine-tuning (Sec.~\ref{sec:ablations}). Similarly, for the real world, multi-step inference confers no consistent advantage in terms of success rate or trajectory smoothness (Sec.~\ref{subsec:sr-analysis} and \ref{subsec:jerk-analysis}).
    \item \textbf{Longer task horizons expose greater sensitivity to inference steps:} The two smaller models (FLOWER, SmolVLA) retain a performance gap that is larger on long-horizon tasks while their short-horizon performance is more robust to the number of integration steps (Sec.~\ref{subsec:sr-analysis}). 
    Importantly, additional Euler steps do not always help: the direction of the performance gap varies across the smaller models (Sec.~\ref{subsec:sr-analysis} and~\ref{sec:ablations}).
    \item \textbf{Practical payoff:} For most current flow matching VLAs, multi-step inference does not improve performance over one step, enabling $1.6$-$4.8\times$ lower latency at deployment (Sec.~\ref{subsec:latency-analysis}).
\end{itemize}

Taken together, these results reframe the conventional wisdom around inference in VLAs. While a growing body of work proposes auxiliary objectives explicitly to recover the single-step regime, we find that for the dominant class of present day flow matching VLAs, the single-step regime is already accessible without any retraining, architectural changes, or additional regularization.

\section{Background and Related Work}
\label{sec:related}
\subsection{Flow matching models}
Flow matching~\cite{lipman2022flow} is a generative modelling framework that learns a time-dependent velocity field $v_\theta(x_t, t)$ defining a continuous-time transport from a source distribution $p_0$ to a target data distribution $p_1$.
The transport is described by the ordinary differential equation (ODE) 
\begin{equation}
    \frac{dx_t}{dt} = v_\theta(x_t, t), \qquad x_0 \sim p_0, t \in [0, 1]
    \label{eq:flow-matching}
\end{equation}
that converts $x_0 \sim p_0$ to $x_1 \sim p_1$.
In rectified flow matching~\cite{liu2022flow}, the driving velocity field is set to follow the direction of the linear path from $x_0$ to $x_1$, i.e., ${v_\theta(x_t, t)=x_1-x_0}$, resulting in ${x_t = (1-t)\,x_0 + t\, x_1}$.
The rectified flow matching objective regresses the predicted velocity onto this target:
\begin{equation}
    \mathcal{L}_{\text{FM}}(\theta) = \mathbb{E}_{t,\, x_0,\, x_1} \big\| v_\theta(x_t, t) - (x_1 - x_0) \big\|_2^2, \qquad \text{with } t \sim \rho
    \label{eq:rectified-flow-matching}
\end{equation}
where $\rho$ is a distribution in $[0,1]$. As discussed in~\cite{liu2022flow}, the minimiser of Eq.~\ref{eq:rectified-flow-matching} is $v^\star(x,t)=\mathbb{E}[x_1-x_0|x_t=x]$ and it is straight iff the interpolation paths do not cross.

To generate a sample from the target distribution, the learned velocity field is integrated from $t=0$ to $t=1$. 
The simplest and most used integrator for flow matching is the explicit Euler scheme with $N$ uniform steps of size $h = 1/N$:
\begin{equation}
    x_{t + h} = x_t + h \cdot v_\theta(x_t, t) \quad t=0,1/N,\dots,(N-1)/N \ .
    \label{eq:euler-integration}
\end{equation}
The choice of $N$ trades inference cost against discretization error, and the choice of a single Euler step ($N=1$) is exact only if the flow is straight. The reference $N$ reported by published VLAs ranges from $4$ to $10$ (Tab.~\ref{tab:model_overview}). In this work, we ask whether using $N=1$ leads to a drop in performance for robot action generation.
\subsection{Accelerating flow and diffusion inference}

The cost of iterative sampling has motivated a broad line of work on reducing the number of integration steps required at inference time.
Consistency models~\cite{kim2024consistency}, shortcut models~\cite{frans2025one}, and MeanFlow~\cite{geng2026mean} augment the training objective to recover the multi-step solution from a single forward pass.
The motivation for doing so is well-documented in image generation, where the quality of the generated image shows that vanilla flow and diffusion models degrade sharply when collapsed into a few Euler steps~\cite{shaul2024bespoke}, which further motivates work on improved ODE solvers and sampling schedules~\cite{song2020denoising, lu2022dpm}.


The few-step methods that have been applied to robotics, MeanFlow~\cite{zhan2026mean} and ManiFlow~\cite{yan2025maniflow0}, were evaluated on small-scale policies where empirically vanilla flow matching policies with single-step inference perform substantially worse than with multi-step inference. Recent work~\cite{pan2026much} shows that the choice of the model architecture is a major component of policy performance. This motivates asking whether the one-step degradation actually remains with large flow matching VLAs and, to the best of our knowledge, this has not been thoroughly investigated.

\section{Experimental Design}
\label{sec:exp-design}

Our central question is whether the multi-step Euler integration is buying meaningful improvements in task success rates over a single Euler step. We answer this with a paired comparison protocol. For each (model, benchmark) pair, we take a single trained policy, and evaluate once with the reference number of integration steps reported by the original paper, and once with a single Euler step.  To reduce stochasticity in the evaluation process, for every (model, benchmark) pair, the 1-step and reference-step rollouts have the same initial conditions \cite{barreiros2026careful}. In simulation this means matched initial seeds for each episode trial. For real world evaluation, trials are run as paired A/B tests. To further control for experimenter bias, the experimenter is blind to the variant being tested.
Finally, in real world experiments we also seed the noise sampled from the policy before every rollout within the pair.

\subsection{Models}

We evaluate six flow matching VLAs, spanning roughly an order of magnitude in parameter count, and a DiT \cite{jones2025multitaskditpolicy} baseline trained from scratch on each benchmark (Tab.~\ref{tab:model_overview}).

\begin{table*}[!h]
\centering
\caption{
Model overview. We report the total parameter count, VLM backbone details, reference number of Euler integration steps as reported by authors, and action generation latency on a single A100 GPU, with the speedup of 1 step in parentheses.
}
\label{tab:model_overview}
\resizebox{\textwidth}{!}{
\begin{threeparttable}
\begin{tabular}{lrcrccc}
\toprule
Model
& Total params
& VLM backbone
& VLM params
& Reference steps
& \multicolumn{2}{c}{Latency (ms)} \\
\cmidrule(lr){6-7}
& & & & & Reference & 1 Step \\
\midrule

DiT~\cite{jones2025multitaskditpolicy}
& $211$M
& CLIP ViT-B/16 + CLIP text~\cite{radford2021learning}
& $149.2$M
& $10$
& $30$
& $15\,{\scriptscriptstyle (2.0\times)}$ \\

SmolVLA \cite{shukor2025smolvla}
& $450$M
& SmolVLM2~\cite{marafioti2025smolvlm}
& $\sim 350$M
& $10$
& $436$
& $91\,{\scriptscriptstyle (4.8\times)}$ \\

X-VLA \cite{zheng2025xvla}
& $900$M
& Florence-2-L (pruned)~\cite{xiao2024florence}
& $570$M
& $10$
& $215$
& $70\,{\scriptscriptstyle (3.1\times)}$ \\

FLOWER \cite{reuss2025flower}
& $947$M
& Florence-2-L (pruned)~\cite{xiao2024florence}
& $570$M
& $4$--$5$\tnote{1}
& $140$
& $60\,{\scriptscriptstyle (2.3\times)}$ \\

$\pi_{0.5}$ \cite{black2025pi05}
& $3.3$B
& PaliGemma VLM~\cite{beyer2024paligemma}
& $3.0$B
& $10$
& $80$
& $50\,{\scriptscriptstyle (1.6\times)}$ \\

LAP \cite{zha2026lap}
& $3.3$B
& PaliGemma VLM~\cite{beyer2024paligemma}
& $3.0$B
& $10$
& $60$
& $35\,{\scriptscriptstyle (1.7\times)}$ \\

XR-0 \cite{cai2026xiaomi}
& $4.7$B
& Qwen3-VL-4B-Instruct~\cite{bai2025qwen3}
& $\sim 4.0$B
& $5$
& $170$
& $90\,{\scriptscriptstyle (1.9\times)}$ \\

\bottomrule
\end{tabular}
\begin{tablenotes}
  \item[1] 4 steps for simulation and 5 for real world experiments as reported by the model authors.
\end{tablenotes}
\end{threeparttable}
}
\end{table*}

For all simulation benchmarks the VLA checkpoints are obtained from the authors' official releases and evaluated as-is, 
with no extra fine tuning or modification of inference-time parameters (e.g., reference number of Euler steps, chunk length or replanning horizon). For the real world experiments we finetune the models on our internal dataset. For the DiT model, we finetune it for all the benchmarks tested. We report fine tuning details in Sec.\,~\ref{app:real-world-finetuning}, and an analysis over different replanning horizons in Sec.\,~\ref{app:planning-horizons}.
We strive to reuse the released checkpoints and not re-implement models to not introduce any bias, and so that the community can readily reproduce our results and findings.

\subsection{Benchmarks}

We evaluate seven models across four simulation benchmarks and one real world suite.


\textbf{Simulation Benchmarks:}
We test all the models on the \textbf{LIBERO}~\cite{liu2023libero} benchmark.
LIBERO has four task suites (Spatial, Object, Goal, LIBERO-10), each comprising 10 tasks. We report aggregate results over the Spatial, Object and Goal suites (denoted as SGO), and separate results for the long-horizon LIBERO-10 suite.
For evaluation, we run $50$ trials per task, totalling \num{2000} trials per model variant. To test our hypothesis against distribution shift, we also test the models on \textbf{LIBERO+}~\cite{fei2025liberop}, a perturbation-extended version of LIBERO with \num{10030} task instances.
We also evaluate on the \textbf{CALVIN ABC$\rightarrow$D}~\cite{mees2022calvin} setting that tests for zero shot generalization.
CALVIN is a long-horizon language-conditioned benchmark, and evaluation requires policies to complete a sequence of 5 tasks across \num{1000} instruction chains.
We report SR@5 (full-chain success) and average chain length.
Finally, we also include the \textbf{RoboTwin 2.0}~\cite{chen2025robotwin} benchmark,
a $50$-task bimanual manipulation benchmark with two difficulty regimes ("clean" and "randomised").
We use the matched-seed paired protocol (\num{100} episodes per task) from the original benchmark, yielding \num{5000} paired episodes per regime per (model, number of steps) cell. Separately, we also examine how the different models behave on one task from \textbf{D3IL}~\cite{jia2024towards}, a benchmark that explicitly evaluates the model's ability to learn multimodal behaviours.

\begin{figure*}[h]
\centering
    \begin{subfigure}{0.245\textwidth}
        \centering
        \includegraphics[width=\linewidth]{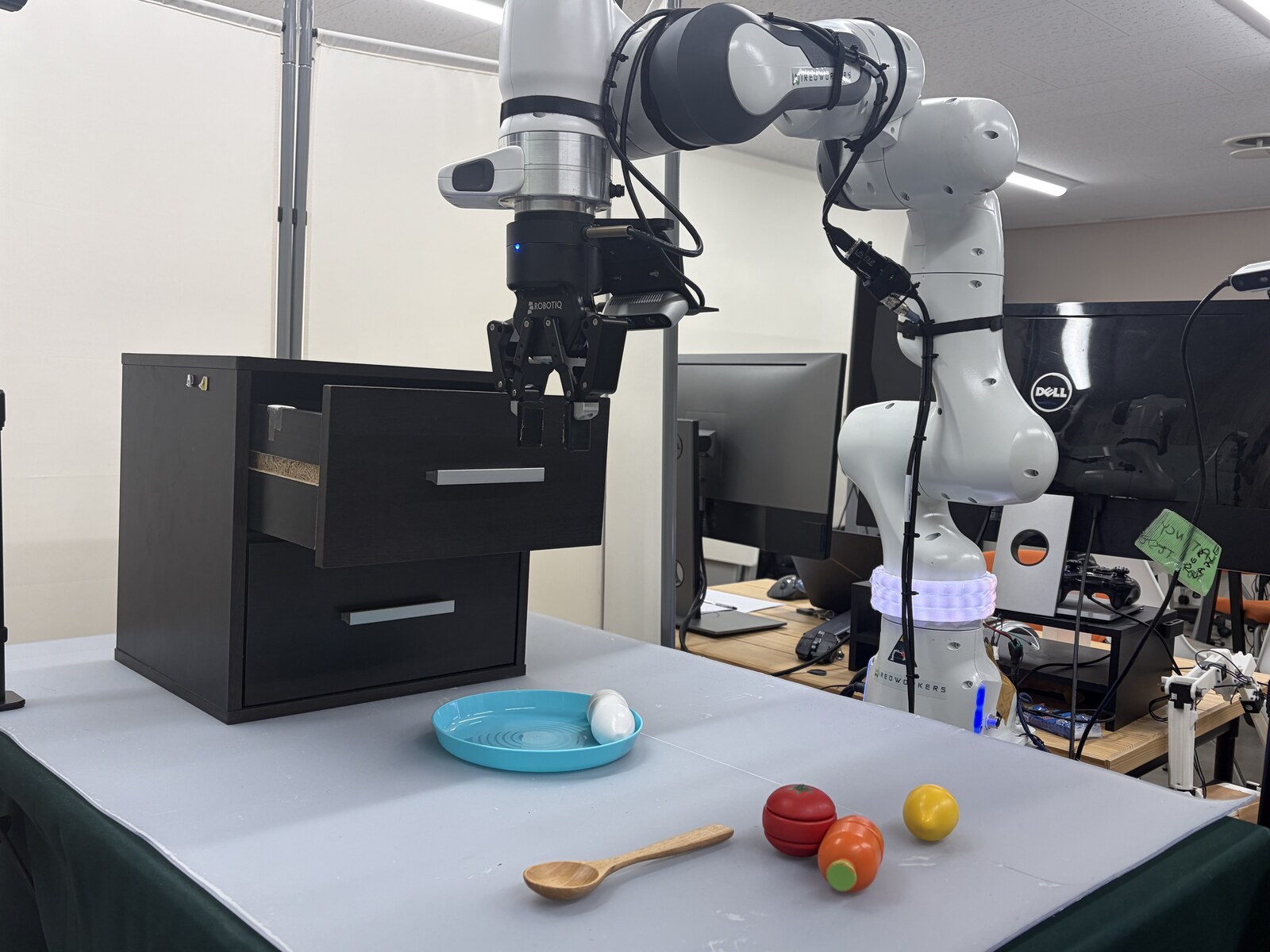}
    \end{subfigure}
    \begin{subfigure}{0.245\textwidth}
        \centering
        \includegraphics[width=\linewidth]{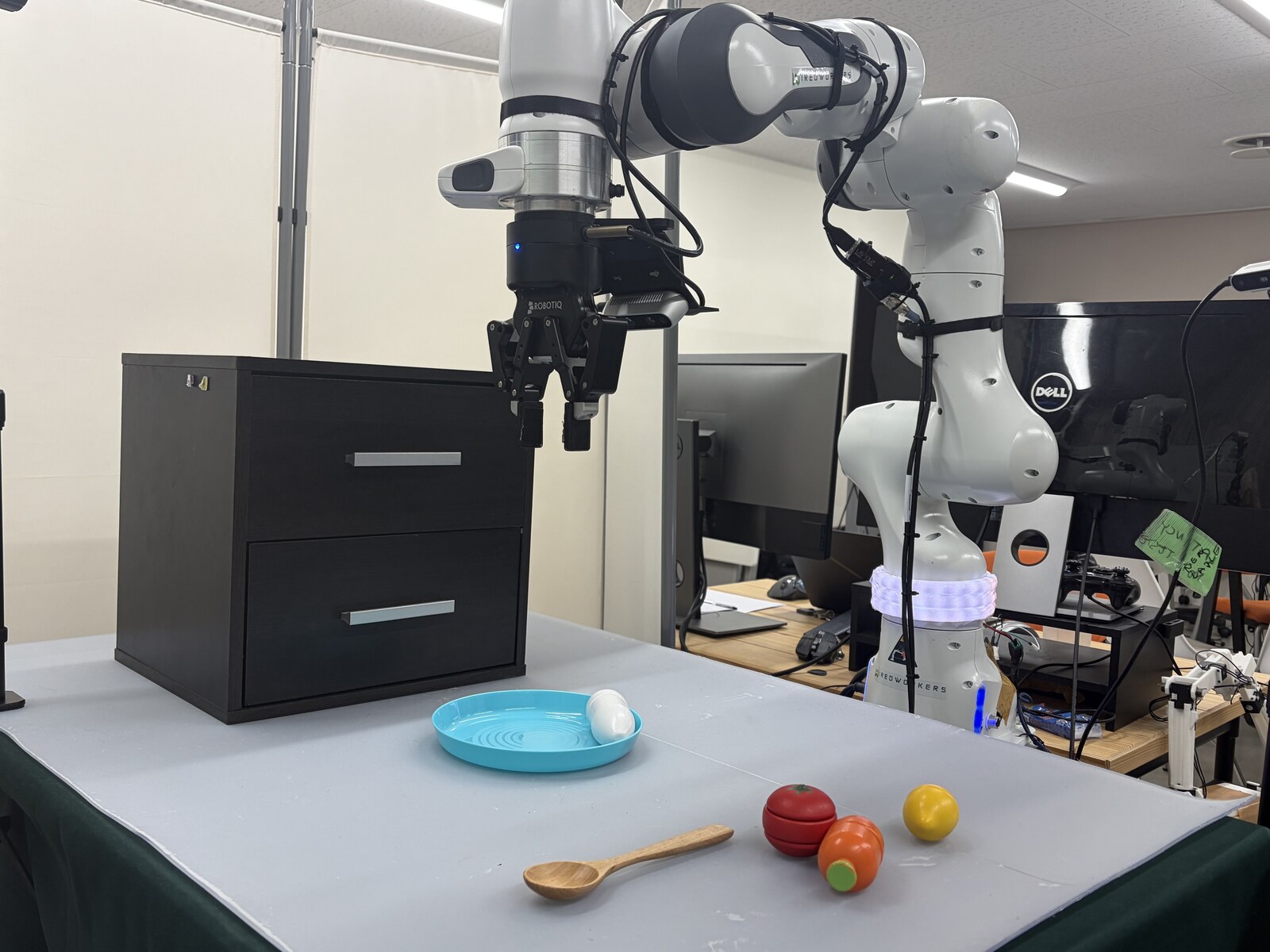}
    \end{subfigure}
    \begin{subfigure}{0.245\textwidth}
        \centering
        \includegraphics[width=\linewidth]{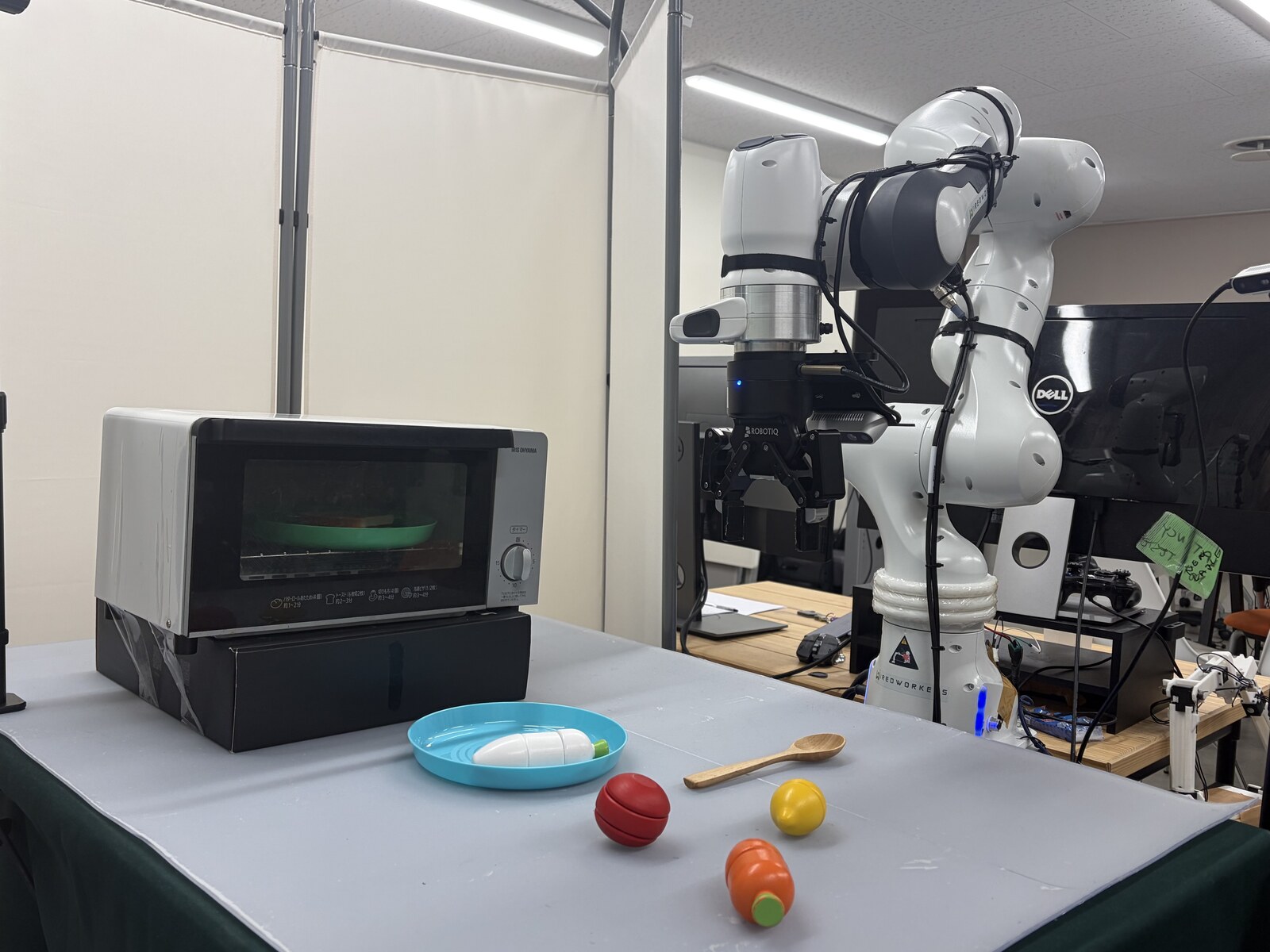}
    \end{subfigure}
    \begin{subfigure}{0.245\textwidth}
        \centering
        \includegraphics[width=\linewidth]{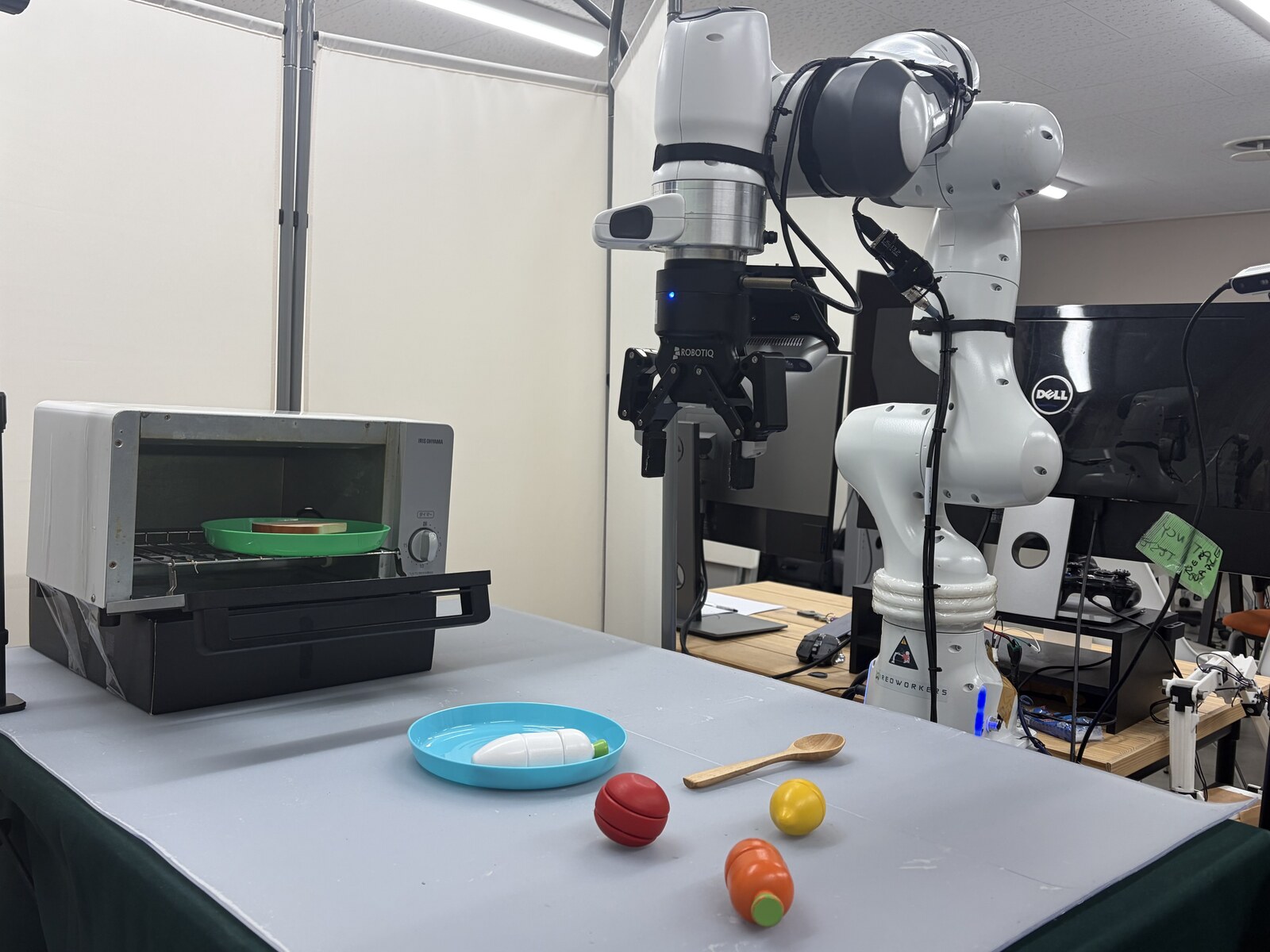}
    \end{subfigure}
    \caption{Real world Franka suite: \textit{spoon drawer}, \textit{open drawer}, \textit{open oven}, \textit{close oven} (left to right).}
    \label{fig:real-world-suite}
\end{figure*}

\textbf{Real world Franka suite:}
This consists of four in-house kitchen-style tasks using a Franka Emika Panda arm 
 (Fig.~\ref{fig:real-world-suite}): \textit{spoon to drawer} (longest-horizon, pick-and-place), \textit{open drawer} (contact-rich), \textit{open oven} (precise manipulation), and \textit{close oven} (short horizon task). The training dataset is the same as the one used in~\cite{zhang2026augmented}. We run $100$ trials per task per (model, num steps), totalling $400$ episodes per model variant and a total of \num{4800} rollouts. Additional evaluation details are provided in the Appendix (Table \ref{tab:benchmark_overview}).

\subsection{Evaluation Protocol}
\label{subsec:eval-pro}

To claim a positive equivalence we test our hypothesis of step-invariance through the TOST procedure~\cite{schuirmann1987comparison} (see Tab.~\ref{tab:tost_all_sim}). We conduct this test aggregating results across tasks within a benchmark: this violates the i.i.d. assumption for the samples of the compared distributions, but the effect is negligible and recommended practice in prior work~\cite{barreiros2026careful}. For real world experiments, due to a more restricted evaluation budget than simulation, we report and discuss the confidence intervals (CIs) of the difference between the approaches, and conduct a McNemar analysis (see Tab.~\ref{tab:realworld-sr} and Sec.~\ref{app:mcnemar}).
We also investigate the quality of the executed real world trajectories through a jerk analysis (see Fig.~\ref{fig:real-world-trajs}). For a per-task breakdown we refer to Sec.~\ref{app:per-task-test}.

For TOST, we set a margin of practical equivalence of $\delta=3\,\textrm{pp}$ (percentage points) when comparing success rates, and $\delta=0.15$ when comparing chain length for CALVIN\footnote{The chain length in CALVIN is up to 5, so we map $3\,$pp to this range} and a confidence level $\alpha=0.05$. To claim equivalence, TOST needs to reject two composite null hypotheses, hence the CIs will be computed at a confidence of $90\%$, i.e., $100(1-2\alpha)\%$. We declare the two methods \colorbox{tostEqBg}{equivalent} within the margin $\delta$ iff the CI for the difference in means of the two distributions is strictly contained in $[-\delta,\delta]$, we declare a \colorbox{tostDiffBg}{relevant difference} if it lies completely outside of it, and otherwise the test is \colorbox{tostIncBg}{inconclusive} (more details in Sec.~\ref{app:appendix-statistics}).

\section{Experimental Results}
\label{sec:results}

We evaluate every (model, benchmark) cell and report results in Tables~\ref{tab:tost_all_sim} and \ref{tab:realworld-sr}: among the 36 test cells in simulation benchmarks 23 show equivalence, 4 show that multi-step is better, 2 show that 1-step is better, and 7 are inconclusive. Step-equivalence appears in models trained on different pretraining corpora, in different codebases, and under different architectural and hyperparameter choices. The presence of this phenomenon across heterogeneous pipelines suggests that it is an effect of well-trained rectified-flow VLAs rather than an artifact of any single model, codebase, or benchmark. The remaining performance gaps are concentrated 1) on long horizon suites, and 2) mainly among $\leq1$B models (DiT, SmolVLA, FLOWER). On short horizon suites and for every model $\geq1$B, single and multi-step inference performance are equivalent based on our tests. A trajectory-smoothness analysis in the real world (Sec.~\ref{subsec:jerk-analysis}) points the same way: multi-step yields smoother motion only for the two sub-1B models, while for every larger model one-step is marginally smoother. Finally, in Sec.~\ref{sec:traj-diversity}, we find that, while not reducing the success rate, one-step inference yields less diverse trajectories.




\subsection{Success Rate Analysis}
\label{subsec:sr-analysis}
\textbf{LIBERO/LIBERO+}. 
The largest VLAs are step-equivalent on the short-horizon SGO suites: X-VLA, $\pi_{0.5}$, LAP, and XR-0 show equivalence and their success rate aggregate differences are within a percentage point, while on LIBERO-10 only LAP shows higher SR with one-step, failing the equivalence test. Non-equivalences come almost entirely from the $\leq1$B models and on the longer-horizon, shifted suites (LIBERO-10, LIBERO+ 10). We further investigate the action difference between the two approaches in Sec.~\ref{app:flow-analysis}.

\begin{table}[t]
\centering
\caption{Simulation metrics with TOST equivalence colouring (Sec.~\ref{subsec:eval-pro}). SGO groups Spatial, Goal and Object suites. -- indicates that no public checkpoint was available for evaluation.}
\label{tab:tost_all_sim}
\resizebox{\linewidth}{!}{%
\begin{tabular}{l*{16}{c}}
\toprule
\multirow{2}{*}{Model}
& \multicolumn{2}{c}{LIBERO SGO}
& \multicolumn{2}{c}{LIBERO-10}
& \multicolumn{2}{c}{LIBERO+ SGO}
& \multicolumn{2}{c}{LIBERO+ 10}
& \multicolumn{2}{c}{CALVIN SR@5}
& \multicolumn{2}{c}{CALVIN Avg. Len.}
& \multicolumn{2}{c}{RoboTwin Clean}
& \multicolumn{2}{c}{RoboTwin Rand.} \\
\cmidrule(lr){2-3}
\cmidrule(lr){4-5}
\cmidrule(lr){6-7}
\cmidrule(lr){8-9}
\cmidrule(lr){10-11}
\cmidrule(lr){12-13}
\cmidrule(lr){14-15}
\cmidrule(lr){16-17}
& Reference & 1 Step
& Reference & 1 Step
& Reference & 1 Step
& Reference & 1 Step
& Reference & 1 Step
& Reference & 1 Step
& Reference & 1 Step
& Reference & 1 Step \\
\midrule

DiT
& \cellcolor{tostIncBg}89.9\%
& \cellcolor{tostIncBg}\textbf{91.9\%}
& \cellcolor{tostDiffBg}\textbf{68.4\%}
& \cellcolor{tostDiffBg}53.8\%
& \cellcolor{tostIncBg}\textbf{30.8\%}
& \cellcolor{tostIncBg}28.6\%
& \cellcolor{tostDiffBg}\textbf{22.0\%}
& \cellcolor{tostDiffBg}14.5\%
& -- & -- & -- & -- & -- & -- & -- & -- \\

SmolVLA
& \cellcolor{tostDiffBg}72.2\%
& \cellcolor{tostDiffBg}\textbf{82.0\%}
& \cellcolor{tostDiffBg}34.6\%
& \cellcolor{tostDiffBg}\textbf{44.2\%}
& \cellcolor{tostEqBg}4.8\%
& \cellcolor{tostEqBg}\textbf{5.0\%}
& \cellcolor{tostEqBg}\textbf{1.2\%}
& \cellcolor{tostEqBg}0.7\%
& -- & -- & -- & -- & -- & -- & -- & -- \\

X-VLA
& \cellcolor{tostEqBg}\textbf{97.6\%}
& \cellcolor{tostEqBg}97.3\%
& \cellcolor{tostEqBg}\textbf{96.2\%}
& \cellcolor{tostEqBg}96.0\%
& \cellcolor{tostEqBg}70.3\%
& \cellcolor{tostEqBg}\textbf{70.4\%}
& \cellcolor{tostEqBg}\textbf{67.4\%}
& \cellcolor{tostEqBg}66.9\%
& \cellcolor{tostIncBg}\textbf{71.0\%}
& \cellcolor{tostIncBg}70.0\%
& \cellcolor{tostEqBg}\textbf{4.18}
& \cellcolor{tostEqBg}4.14
& \cellcolor{tostEqBg}\textbf{65.5\%}
& \cellcolor{tostEqBg}65.1\%
& \cellcolor{tostEqBg}\textbf{24.3\%}
& \cellcolor{tostEqBg}23.3\% \\

FLOWER
& \cellcolor{tostEqBg}\textbf{97.3\%}
& \cellcolor{tostEqBg}96.5\%
& \cellcolor{tostDiffBg}\textbf{89.6\%}
& \cellcolor{tostDiffBg}83.6\%
& \cellcolor{tostIncBg}\textbf{77.6\%}
& \cellcolor{tostIncBg}74.4\%
& \cellcolor{tostDiffBg}\textbf{72.0\%}
& \cellcolor{tostDiffBg}59.3\%
& \cellcolor{tostIncBg}\textbf{78.5\%}
& \cellcolor{tostIncBg}75.5\%
& \cellcolor{tostIncBg}\textbf{4.51}
& \cellcolor{tostIncBg}4.43
& -- & -- & -- & -- \\

$\pi_{0.5}$
& \cellcolor{tostEqBg}98.5\%
& \cellcolor{tostEqBg}\textbf{98.9\%}
& \cellcolor{tostEqBg}94.4\%
& \cellcolor{tostEqBg}94.4\%
& \cellcolor{tostEqBg}\textbf{86.4\%}
& \cellcolor{tostEqBg}86.3\%
& \cellcolor{tostEqBg}\textbf{77.6\%}
& \cellcolor{tostEqBg}76.1\%
& -- & -- & -- & -- & -- & -- & -- & -- \\

LAP
& \cellcolor{tostEqBg}97.7\%
& \cellcolor{tostEqBg}\textbf{98.8\%}
& \cellcolor{tostIncBg}90.2\%
& \cellcolor{tostIncBg}\textbf{92.0\%}
& \cellcolor{tostEqBg}\textbf{77.6\%}
& \cellcolor{tostEqBg}76.9\%
& \cellcolor{tostEqBg}\textbf{66.4\%}
& \cellcolor{tostEqBg}65.7\%
& -- & -- & -- & -- & -- & -- & -- & -- \\

XR-0
& \cellcolor{tostEqBg}\textbf{97.3\%}
& \cellcolor{tostEqBg}97.0\%
& \cellcolor{tostEqBg}96.8\%
& \cellcolor{tostEqBg}\textbf{97.4\%}
& \cellcolor{tostEqBg}75.8\%
& \cellcolor{tostEqBg}\textbf{76.5\%}
& \cellcolor{tostEqBg}\textbf{70.2\%}
& \cellcolor{tostEqBg}69.6\%
& \cellcolor{tostEqBg}87.2\%
& \cellcolor{tostEqBg}\textbf{88.1\%}
& \cellcolor{tostEqBg}4.73
& \cellcolor{tostEqBg}4.73
& -- & -- & -- & -- \\

\bottomrule
\end{tabular}%
}
\end{table}

\begin{table}[t]
\centering
\caption{Real-world success rates (\%) per task and pooled (n=100 per task, 400 pooled). In grey the 90\% CI for $\Delta=p_{\mathrm{ref}}-p_{\mathrm{1}}$. Cells with * marks a significant McNemar difference ($\alpha=0.05$; per-task decisions are after Holm correction within each model; no correction for pooled analysis).}
\label{tab:realworld-sr}
\resizebox{\linewidth}{!}{%
\begin{tabular}{lcccccccccc}
\toprule
\multirow{2}{*}{Model} & \multicolumn{2}{c}{Spoon $\to$ Drawer} & \multicolumn{2}{c}{Open Drawer} & \multicolumn{2}{c}{Open Oven} & \multicolumn{2}{c}{Close Oven} & \multicolumn{2}{c}{Pooled} \\
\cmidrule(lr){2-3} \cmidrule(lr){4-5} \cmidrule(lr){6-7} \cmidrule(lr){8-9} \cmidrule(lr){10-11}
 & Reference & 1 Step & Reference & 1 Step & Reference & 1 Step & Reference & 1 Step & Reference & 1 Step \\
\midrule
\multirow{2}{*}{DiT}
& \textbf{33.0} & 32.0 & 59.0 & \textbf{64.0} & 0.0 & \textbf{1.0} & 22.0 & \textbf{23.0} & 28.5 & \textbf{30.0} \\[-1pt]
& \multicolumn{2}{c}{\textcolor{black!55}{\scriptsize $[-8.2,\,+10.2]$}} & \multicolumn{2}{c}{\textcolor{black!55}{\scriptsize $[-13.5,\,+3.6]$}} & \multicolumn{2}{c}{\textcolor{black!55}{\scriptsize $[-4.4,\,+1.7]$}} & \multicolumn{2}{c}{\textcolor{black!55}{\scriptsize $[-7.2,\,+5.2]$}} & \multicolumn{2}{c}{\textcolor{black!55}{\scriptsize $[-5.0,\,+2.0]$}} \\[3pt]
\multirow{2}{*}{X-VLA}
& 0.0 & 0.0 & 68.0 & \textbf{72.0} & 0.0 & 0.0 & 49.0 & \textbf{50.0} & 29.2 & \textbf{30.5} \\[-1pt]
& \multicolumn{2}{c}{\textcolor{black!55}{\scriptsize $[-2.6,\,+2.6]$}} & \multicolumn{2}{c}{\textcolor{black!55}{\scriptsize $[-9.5,\,+1.4]$}} & \multicolumn{2}{c}{\textcolor{black!55}{\scriptsize $[-2.6,\,+2.6]$}} & \multicolumn{2}{c}{\textcolor{black!55}{\scriptsize $[-7.5,\,+5.5]$}} & \multicolumn{2}{c}{\textcolor{black!55}{\scriptsize $[-3.4,\,+0.9]$}} \\[3pt]
\multirow{2}{*}{FLOWER}
& 46.0 & \textbf{50.0} & 81.0 & \textbf{97.0}\textsuperscript{*} & 20.0 & \textbf{31.0} & \textbf{92.0} & 75.0\textsuperscript{*} & 59.8 & \textbf{63.2} \\[-1pt]
& \multicolumn{2}{c}{\textcolor{black!55}{\scriptsize $[-13.8,\,+5.9]$}} & \multicolumn{2}{c}{\textcolor{black!55}{\scriptsize $[-22.9,\,-10.0]$}} & \multicolumn{2}{c}{\textcolor{black!55}{\scriptsize $[-20.0,\,-1.8]$}} & \multicolumn{2}{c}{\textcolor{black!55}{\scriptsize $[+9.4,\,+24.8]$}} & \multicolumn{2}{c}{\textcolor{black!55}{\scriptsize $[-7.7,\,+0.7]$}} \\[3pt]
\multirow{2}{*}{$\pi_{0.5}$}
& 88.0 & \textbf{90.0} & \textbf{78.0} & 77.0 & 63.0 & \textbf{73.0} & \textbf{87.0} & 74.0\textsuperscript{*} & \textbf{79.0} & 78.5 \\[-1pt]
& \multicolumn{2}{c}{\textcolor{black!55}{\scriptsize $[-7.8,\,+3.7]$}} & \multicolumn{2}{c}{\textcolor{black!55}{\scriptsize $[-3.8,\,+5.8]$}} & \multicolumn{2}{c}{\textcolor{black!55}{\scriptsize $[-19.7,\,-0.1]$}} & \multicolumn{2}{c}{\textcolor{black!55}{\scriptsize $[+6.4,\,+19.9]$}} & \multicolumn{2}{c}{\textcolor{black!55}{\scriptsize $[-3.0,\,+4.0]$}} \\[3pt]
\multirow{2}{*}{LAP}
& \textbf{75.0} & 73.0 & \textbf{78.0} & 76.0 & \textbf{56.0} & 47.0 & \textbf{88.0} & 86.0 & \textbf{74.2} & 70.5 \\[-1pt]
& \multicolumn{2}{c}{\textcolor{black!55}{\scriptsize $[-4.8,\,+8.8]$}} & \multicolumn{2}{c}{\textcolor{black!55}{\scriptsize $[-3.0,\,+7.1]$}} & \multicolumn{2}{c}{\textcolor{black!55}{\scriptsize $[+1.1,\,+16.7]$}} & \multicolumn{2}{c}{\textcolor{black!55}{\scriptsize $[-2.6,\,+6.8]$}} & \multicolumn{2}{c}{\textcolor{black!55}{\scriptsize $[+0.7,\,+6.8]$}} \\[3pt]
\multirow{2}{*}{XR-0}
& \textbf{73.0} & 69.0 & 81.0 & \textbf{83.0} & 42.0 & 42.0 & 88.0 & 88.0 & \textbf{71.0} & 70.5 \\[-1pt]
& \multicolumn{2}{c}{\textcolor{black!55}{\scriptsize $[-3.1,\,+11.1]$}} & \multicolumn{2}{c}{\textcolor{black!55}{\scriptsize $[-6.7,\,+2.5]$}} & \multicolumn{2}{c}{\textcolor{black!55}{\scriptsize $[-9.6,\,+9.6]$}} & \multicolumn{2}{c}{\textcolor{black!55}{\scriptsize $[-5.3,\,+5.3]$}} & \multicolumn{2}{c}{\textcolor{black!55}{\scriptsize $[-2.9,\,+3.9]$}} \\[3pt]
\bottomrule
\end{tabular}
}
\end{table}

\textbf{CALVIN ABC$\rightarrow$D}. 
The largest VLA model is step-invariant on the long-horizon, zero-shot transfer setting while the smaller models exhibit degradation in performance with one-step inference. 
For the two smaller models, the CI of the SR gap is $[+0.4, +4.6]$ for FLOWER and $[-1.0, +3.0]$ for X-VLA. FLOWER's CI does not include 0 so there is a statistical difference in performance between the two inference settings, while for X-VLA the CI reaches the margin of our equivalence bounds making the test inconclusive (more details in Sec. \ref{app:appendix-statistics-calvin}).

\textbf{RoboTwin}. 
X-VLA, the only model with a public checkpoint on this benchmark, lands within a percentage point on both the clean and randomised suites, and TOST returns an equivalence verdict. This extends our claim to a bimanual manipulation setting, although with a single model.

\textbf{Real World} (Table ~\ref{tab:realworld-sr}). Unlike in simulation, our physical rollouts cannot resolve the pre-registered equivalence at this evaluation budget (see the CI for the degenerate X-VLA cells). In simulation, it is possible to run many rollouts cheaply, and fixed seeds with near deterministic dynamics make the coupling much tighter. For real robot evaluation these factors are not there, inflating the variance of the estimated gap. 
To avoid resting on absence of evidence, we also report the results of a McNemar test (see Sec.~\ref{app:mcnemar}). Only 3 of 24 cells exhibit a genuine difference and they are bidirectional: the multi-step reference is better on \emph{Close Oven} for $\pi_{0.5}$ (${p=2.3\textrm{e-3}}$) and FLOWER (${p=5\textrm{e-4}}$), while 1-step is better on FLOWER \emph{Open Drawer} (${p=3.1\textrm{e-5}}$). Pooled per-model gaps are uniformly small and in both directions ($|\Delta| \leq 3.8\,$pp; 4 of 6 models within $\pm2\,$pp). LAP is the only model where the CI of the pooled difference does not straddle 0, and on the pooled results the McNemar test results in \(p=0.0545\), so a genuine multi-step advantage cannot be excluded for this model. Therefore, our real world results indicate that multi-step inference confers no systematic or consistent advantage over 1-step inference in the real world, rather than asserting equivalence as in simulation.

\subsection{Inference Latency}
\label{subsec:latency-analysis}

On a single A100 GPU, collapsing to 1-step cuts latency by $1.6\times$ to $4.8\times$, raising the achievable control frequency (see Tab.~\ref{tab:model_overview}). Since only the action head re-runs per Euler step, models with a small action head relative to the VLM gain the least ($\pi_{0.5}$, $1.6\times$), while those where the action head is comparable gain the most (SmolVLA, $4.8\times$).
For the larger models, the latency gains generally come at little cost to task success, with the exceptions noted above.

\subsection{Trajectory smoothness}
\label{subsec:jerk-analysis}

\begin{figure}[h]
\centering
    \includegraphics[width=\linewidth]{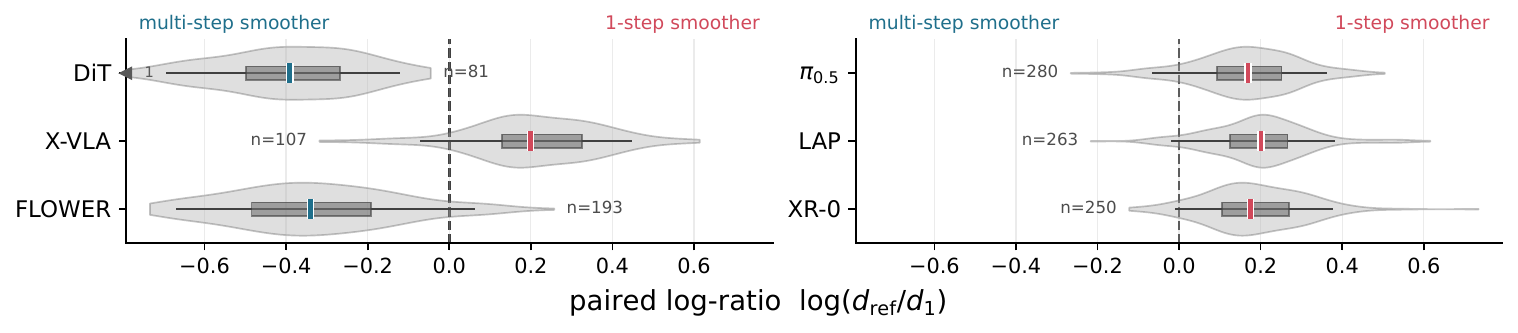}
    \caption{Paired jerk-RMS comparison of real world rollouts. Violins are kernel density estimates; boxes give quartiles with 5–95\% whiskers; the coloured line is the median. $n$ is the number of samples (pairs with both runs successful).}
    \label{fig:real-world-trajs}
\end{figure}

Beyond success rates, we ask whether one-step inference changes the smoothness of the executed motion. To quantify trajectory smoothness we use the per-episode root mean square (RMS) of the joint jerk: after recording the executed joint trajectory, we first smooth it with a Savitzky-Golay filter (polynomial of order \num[]{3}, window size \num[]{11}) and then compute the jerk via finite differences.
Because failure modes can produce different jerk profiles, we only compare episodes where both approaches succeed. We report the per-model distribution of the log ratio of the paired joint-space jerk in Fig.~\ref{fig:real-world-trajs}. We find no consistent smoothness cost to one-step inference: the multi step reference is meaningfully smoother only for two sub-1B models (DiT, FLOWER), while for every model above 1B, one-step is in fact marginally smoother, echoing the scale-dependence of our task-success results.

\subsection{Trajectory diversity}
\label{sec:traj-diversity}
\begin{figure}[h]
\centering
    \begin{subfigure}{0.65\textwidth}
        \centering
        \includegraphics[width=\linewidth]{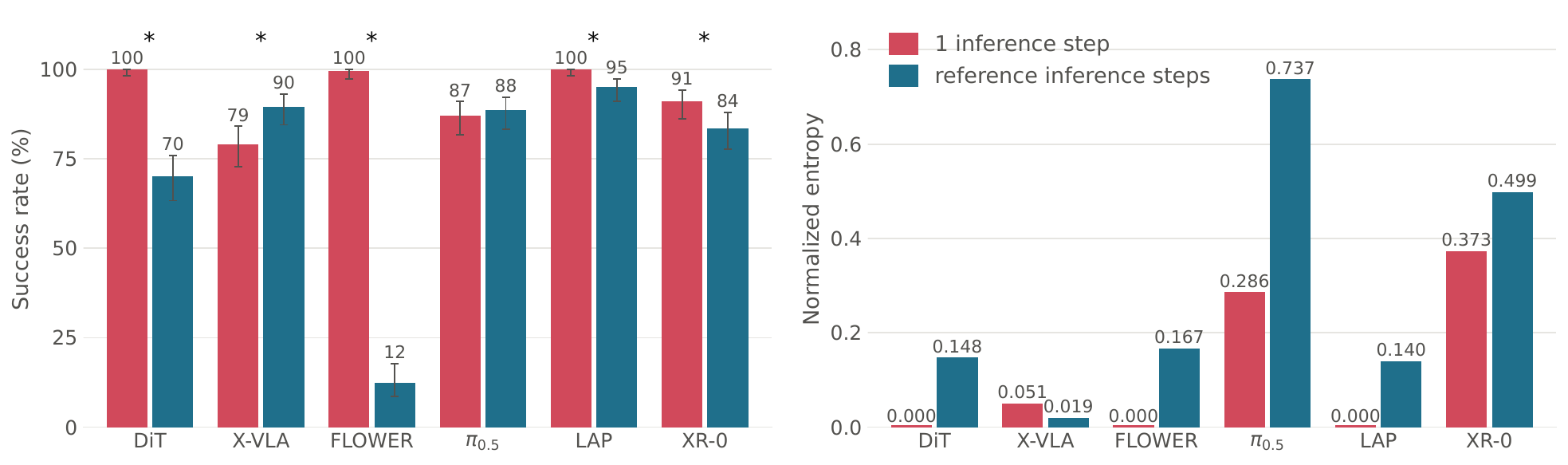}
        \caption{Success rate with 95\% CIs and normalised entropy}
        \label{fig:d3il-bars}
    \end{subfigure}
    \begin{subfigure}{0.33\textwidth}
        \centering
        \includegraphics[width=\linewidth]{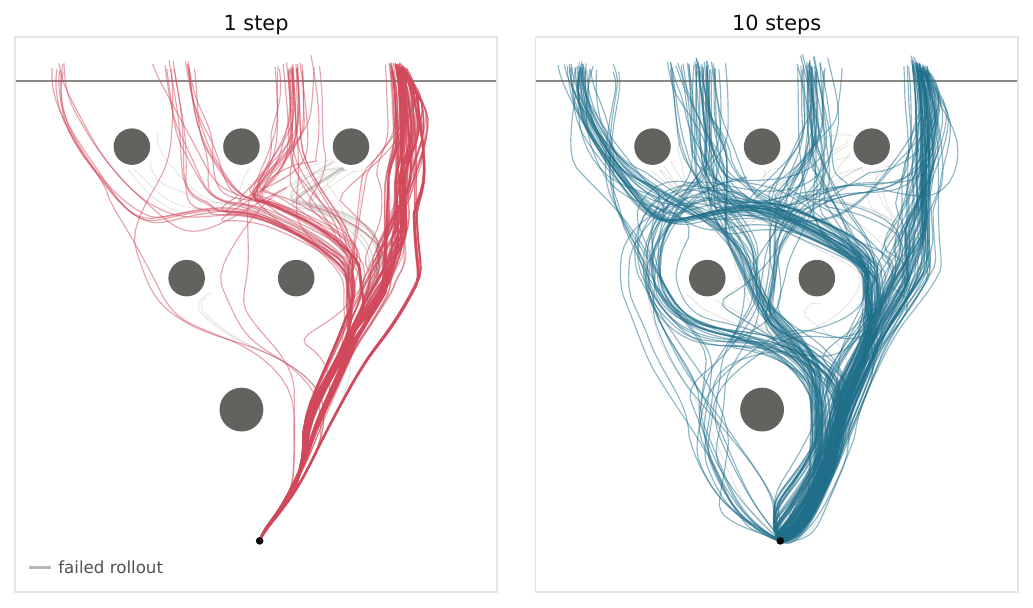}
        \caption{Trajectories executed by $\pi_{0.5}$}
        \label{fig:d3il-paths}
    \end{subfigure}
    \caption{D3IL Avoiding results. In (a), $^*$ indicates a significant McNemar difference (${p < 0.05}$).}
    \label{fig:d3il}
\end{figure}

We evaluated the models on D3IL Avoiding \cite{jia2024towards}, a multimodal control task with 24 valid success modes, to study policy behaviour diversity. As the original benchmark only has a state-based version of the Avoiding task, we built a visual version of this dataset by collecting image data replaying the state demos. We evaluated each model through 200 evaluation episodes on matched initial state and policy noise stream. Fine-tuning details can be found in Tab.~\ref{tab:deploy_train_hparams} and Sec.~\ref{app:appendix-d3il}.

As shown in Fig.~\ref{fig:d3il-bars}, one-step inference improves performance for DiT and FLOWER, degrades it for X-VLA, while among larger models one-step inference consistently does not hurt success rate, even increasing it for LAP and XR-0. However, it changes behaviour diversity by generating trajectories with less entropy, i.e., concentrating the generated trajectories on fewer valid strategies. Under one-step inference, DiT, FLOWER, and LAP each produce a single successful strategy, yielding zero strategy entropy, whilst maintaining success rates of at least 99.5\%. Multi-step inference recovers multiple successful strategies, although with lower overall success. For the $\pi_{0.5}$ model, while the two approaches have similar success rates, one-step inference has far less entropy, and the number of observed successful strategies drops from 19 to 8 (Fig.~\ref{fig:d3il-paths}). Together, these findings suggest that one-step inference does not hurt success rate, but does narrow the range of generated behaviours.

\subsection{Ablations}
\label{sec:ablations}

\begin{figure*}[htp]
\centering
  \centering
  \includegraphics[width=\linewidth]{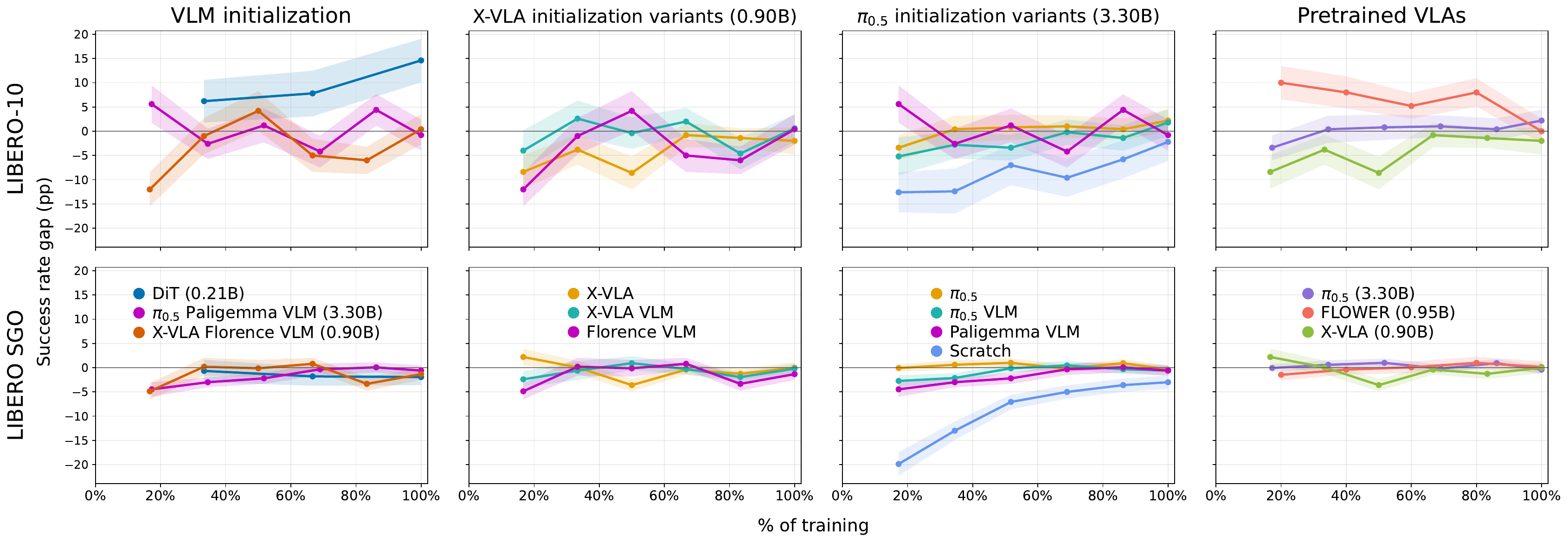}
  \caption{Step-count gap over training on LIBERO. Each panel reports $\Delta=\text{SR}_\text{ref}-\text{SR}_{1}$ with shaded area as 90\% CIs. Multi-step gains, when present, are concentrated on LIBERO-10 and fade with training. The short-horizon suites are close to step-invariant throughout.}
  \label{fig:ablation_gap_over_training}
\end{figure*}

The main results show that multi-step Euler inference does not guarantee higher success rates for current flow matching VLAs. We now investigate two questions: (Q1) How do model scale and pretraining influence step-equivalence? (Q2) How does the gap evolve over the course of training? 
We focus on LIBERO because it contains both short-horizon SGO suites, and the longer-horizon LIBERO-10 suite, where most of the performance gaps appear. Our findings are visualised in Fig.~\ref{fig:ablation_gap_over_training}.

\textbf{Q1: How do model scale and pretraining influence step-equivalence?}
We first consider the randomly initialised $\pi_{0.5}$ (scratch; Fig.~\ref{fig:ablation_gap_over_training}, third column). It favours one-step inference throughout training and has not reached step-equivalence by the final checkpoint, although its gap narrows substantially (from $-20$ to $-3$\,pp on SGO with a $[-4.38, -1.64]$ CI). 
In the second column of Fig.~\ref{fig:ablation_gap_over_training}, we omit the from-scratch X-VLA variant because of its low absolute task performance: across evaluated checkpoints and inference-step settings, success rates remain below 10\% on LIBERO-10, 5\% on Goal, and 20\% on Spatial, and peak at 42.8\% on Object. Despite its low absolute performance, from-scratch X-VLA favours one-step inference more often, including on pooled SGO at the final two checkpoints, where the 90\% CIs for the performance gap lie entirely below zero.
Thus, within our training budget, step-equivalence does not emerge from random initialization, though the converging trend for $\pi_{0.5}$ suggests that longer training might close the gap.

To identify which kind of pretraining matters, we compare different initializations for both $\pi_{0.5}$ and X-VLA: a generic VLM (PaliGemma2~\cite{beyer2024paligemma} or Florence~\cite{xiao2024florence}, respectively) with a randomly initialised action expert, the VLA's own VLM backbone (trained also on robot data) with a randomly initialised action expert, and the fully pretrained VLA. For both models, all three achieve step-equivalence on SGO and reach near-zero gaps by the final checkpoint, so robot-action pretraining is not required. However, pretrained vision-language features alone are not sufficient: the 211M DiT, which conditions on pretrained CLIP image and text encoders~\cite{radford2021learning} but trains the remaining model from scratch, develops a multi-step advantage on LIBERO-10 that grows to $15\,$pp over training (first column), echoing our scale-dependent main results.

\textbf{Q2: How does the gap evolve over the course of training?} $\pi_{0.5}$ is step-equivalent on SGO from the earliest checkpoint. For X-VLA, early in training, additional Euler steps hurt performance, and this gap collapses by the final checkpoint. FLOWER is the clearest case where multi-step helps, but the benefit is localised to LIBERO-10 at early and intermediate checkpoints, while the short-horizon SGO gap remains near zero and step-equivalence is verified on both SGO and LIBERO-10 at the last checkpoint (Fig.~\ref{fig:ablation_gap_over_training}, rightmost column).

This last point resolves an apparent tension with Tab.~\ref{tab:tost_all_sim} where FLOWER shows a residual gap on LIBERO while X-VLA, a model of similar scale ($950$M), does not. The discrepancy is one of checkpoint selection, not scale. The authors of FLOWER train their model for up to 40 epochs but mention that they use early stopping and for LIBERO use the checkpoint at epoch 19. Our ablation sweep instead retains all checkpoints during training, and at the final checkpoint FLOWER's step-count performance gap disappears, bringing it in line with X-VLA. The Tab.~\ref{tab:tost_all_sim} gap therefore reflects an early-stopped checkpoint evaluated before step-invariance has fully emerged, rather than a property of FLOWER's scale or architecture.

Together, these refine the main claim: step-equivalence is favoured by larger models, generic VLM pretraining, 
and longer fine-tuning. When one of these properties is missing, tasks with longer horizons tend to show step-dependence.

\section{Limitations}
\label{sec:limitations}

To aid reproducibility and avoid confounding step-invariance with our own training decisions, we deliberately used released checkpoints as-is rather than fine-tuning base models for all our evaluations in simulation. 
The downside is uneven coverage: not every model ships a checkpoint for every benchmark, so some (model, benchmark) cells could not be evaluated.
A full grid would require training each model on each benchmark, which we leave to future work.
While the aggregate results for the real world tasks are close, we observed up to $13$ pp difference between single- and multi-step regimes on single tasks.
Real world evaluation could therefore benefit from an increased dataset size, matching the scale of the simulated benchmarks.
Finally, we show the single-step policy matches the multi-step reference in most cases and that single-step gradient computations are therefore cheap. However, we do not run RL experiments to validate that resulting policy gradient updates are stable and effective at VLA scale.

\section{Conclusion}
\label{sec:conclusion}
Across six flow matching VLAs from 450M to 4.7B parameters, plus a $211\,$M DiT, four simulation benchmarks and a real world Franka suite, we find the assumption that multi-step inference is necessary for task performance does not hold for well-trained large VLAs. Where multi-step helps, the benefit is localised to smaller models on long-horizon tasks. Our ablations trace step-invariance to a combination of large scale models, training maturity and the quality of the pretrained VLM features rather than to large-scale robot-action pretraining which we find to be not necessary.

The key implication is deployment latency: $N$ multiplies action-head compute at every replanning step, limiting control frequency. A single Euler step removes this overhead with negligible impact on success rate, enabling low-latency deployment of existing flow-matching VLA checkpoints without retraining, distillation, or auxiliary objectives.
The same property makes single-backward-pass actor updates attractive for RL post-training without differentiating through the integration chain. Converting this into a concrete recipe is a promising direction for future work.

\acknowledgments{This work was supported by JST under Moonshot R\&D Grant Number JPMJMS2012. This work was also partially supported by a Lambda, Inc. research grant. We also thank the reviewers for the discussion and feedback during the rebuttal period.}

\bibliography{main}
\newpage
\appendix
\renewcommand{\thefigure}{\thesection\arabic{figure}}
\renewcommand{\thetable}{\thesection\arabic{table}}
\renewcommand{\theequation}{\thesection\arabic{equation}}

\makeatletter
\@addtoreset{figure}{section}
\@addtoreset{table}{section}
\@addtoreset{equation}{section}
\makeatother

\addcontentsline{toc}{section}{Appendix} 
\part{Appendix} 
\mtcsetdepth{parttoc}{3} 
\parttoc 
\newpage



\section{Action space disagreement and flow geometry}
\label{app:flow-analysis}
Our main results show that one-step and multi-step Euler inference often achieve statistically equivalent success rates. However, as shown in Sec.~\ref{sec:traj-diversity}, similar task performance does not necessarily imply that the two methods produce identical actions. 
We investigate this distinction by examining how action-space disagreement relates to task success during training in Sec.~\ref{app:appendix-action-space}, and whether the geometry of the learned flow explains the benefits of additional integration steps~\ref{app:flowgeometry}.

\subsection{Action space divergence analysis}
\label{app:appendix-action-space}

\begin{figure*}[htp]
\centering
    \includegraphics[width=\linewidth]{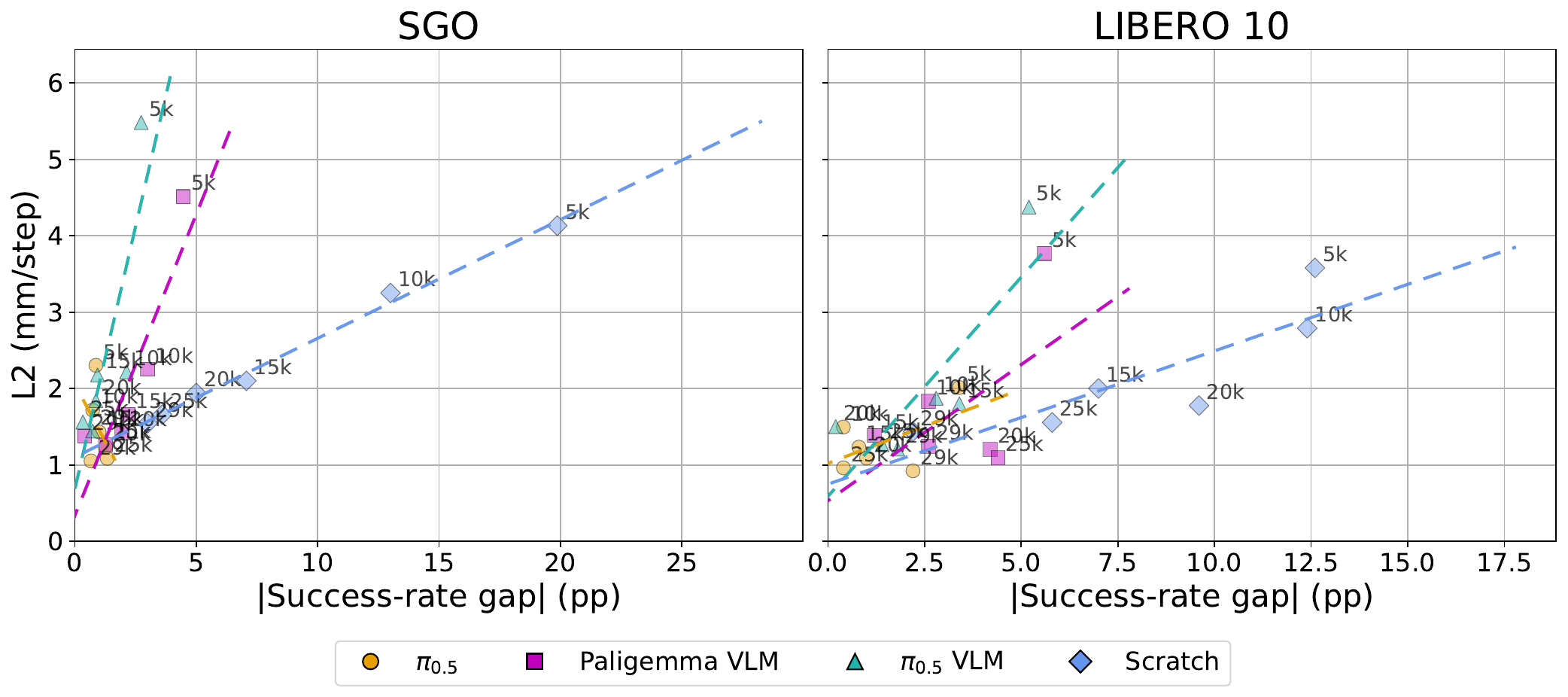}
    \caption{Comparison between the 1-step and 10-step integration approaches: L2 distance in the action space against the absolute value of the success rate gap. Each marker is a checkpoint (labeled 5k–29k), and color/shape of the marker identify the initialization variant. Dashed lines capture per-variant trend.
}
    \label{fig:l2-sr-plots}
\end{figure*}
In our ablations (Sec.~\ref{sec:ablations}) we investigated how the success rate (SR) gap between the 1-step and multi-step approach varies during training. To further investigate the mechanism behind this gap, we here analyse the action divergence, i.e., the L2 distance between the 1-step and reference-step actions, which corresponds to 10 steps for the $\pi_{0.5}$ model family.
We chose to conduct this analysis on the LIBERO benchmark because it contains both short-horizon Spatial, Goal, Object (SGO) suites, and the longer-horizon LIBERO-10 suite.

We consider the same four initializations as in Sec.~\ref{sec:ablations}: 1) the base \textcolor{orange}{$\bm{\pi_{0.5}}$} model, 2) \textcolor{lightseagreen}{$\bm{\pi_{0.5}}$\textbf{VLM}} weights with a randomly initialised action expert, 3) \textcolor{magenta}{\textbf{Paligemma VLM}} initialised with Paligemma2 weights~\cite{beyer2024paligemma}, and 4) \textcolor{cornflowerblue}{\textbf{Scratch}}, where all weights are randomly initialised. All the models are finetuned for \num{30000} gradient steps and with a batch size of 256.

We randomly sample $5,000$ transitions for each LIBERO suite, and measure the L2 distance between actions generated with 1 and 10 Euler steps, using the same initial noise\footnote{We report the L2 distance over only the first 3 action dimensions so that the distance has interpretable units (\SI{}{\mm}/step). Note that an equivalent plot considering the rotation dimensions (\SI{}{\deg}/step) shows the same trend.} against the SR gap.

Fig.~\ref{fig:l2-sr-plots} shows that the two quantities shrink together over training as every variant moves toward the origin, but the correlation between the observed gap and the action divergence decreases as the amount of pretraining increases. The pretrained variants \textcolor{lightseagreen}{$\bm{\pi_{0.5}}$ \textbf{VLM}} and \textcolor{magenta}{\textbf{Paligemma VLM}} can have large action divergence early in training yet exhibit an SR gap of less than five percentage points, while the \textcolor{cornflowerblue}{\textbf{Scratch}} variant turns a slightly lower L2 distance into an $18$ percentage point gap.

Thus, the magnitude of action disagreement alone does not determine its effect on task success.
We hypothesise that this decoupling reflects multi-modality in the action distribution: with informative VLM features, the 1-step and 10-step solutions can land on different but equally valid action modes, so a large L2 difference need not imply a worse outcome.
Without such features, as in the randomly initialised variant, the 1-step approach instead drifts toward genuinely worse actions.

\subsection{Flow geometry and endpoint disagreement}
\label{app:flowgeometry}
We next investigate whether the geometry of the learned flow can explain when additional Euler steps improve performance. Given a noise sample $x_0 \sim \mathcal{N}(0, I)$, we indicate with $a_1$ and $a_N$ the action obtained with one-step and $N$ step inference respectively. We define the chord as $c = \|a_N - x_0\|$, that is the distance the flow transports.

We study the geometry of the learned flow field across models through two normalised metrics
\begin{itemize}
    \item \textbf{Geometric excess} (\texttt{Excess}): we measure the sum of the norm of the instantaneous velocities divided by the chord ${\frac{\sum_{i=0}^{N-1} \|x_{(i+1)/N} - x_{i/N}\|}{c}}\,$. It is the length of the integrated path relative to the straight line between its endpoints. It is exactly 1 when every step points in the same direction, and its excess over 1 grows with the total turning of the path. We report the excess over $1$ as a percentage for easier reading.
    \item \textbf{Endpoint error} (\texttt{Error}): we report the disagreement between the one-step and multi-step action divided by the chord ${\frac{\|a_1 - a_{\mathrm{ref}}\|}{c}}\,$. 
\end{itemize}


We evaluate both metrics on $4000$ LIBERO demonstration observations ($1000$ per suite) with eight fixed noise draws per observation. We compute the the mean of both metrics over the eight draws, and we report the mean over observations. Six of the seven checkpoints travel within 0.7\% of their chord (Tab.~\ref{tab:app-flow-released}), including every step-invariant model. The exception is SmolVLA with an excess of $2.35\%$.
\begin{table}[h]
  \centering
  \small
  \begin{tabular}{l r r rr}
    \toprule
    Model & Params & $k$ & \texttt{Excess} (\%) & \texttt{Error}\ (\%) \\
    \midrule
    DiT         & 211M & 10 &  0.59   & 9.87 \\
    SmolVLA     & 450M & 10 &  2.35   & 29.29 \\
    X-VLA       & 900M & 10 & $<$0.01 &  0.52 \\
    FLOWER      & 947M &  4 &  0.24   &  4.56 \\
    $\pi_{0.5}$ & 3.3B & 10 &  0.25   &  4.85 \\
    LAP         & 3.3B & 10 &  0.27   &  5.00 \\
    XR-0        & 4.7B &  5 &  0.64   &  8.97 \\
    \bottomrule
  \end{tabular}
  \caption{Flow geometry and normalised endpoint error across released VLA checkpoints, evaluated on $4000$ LIBERO demonstration observations ($1000$ per suite), with eight fixed noise draws per observation. \texttt{Excess} measures the relative length of the integration path beyond the straight-line distance between its endpoints; \texttt{Error} measures normalised disagreement between one-step and reference-step actions. Smaller values indicate straighter trajectories and closer action agreement, respectively.}
  \label{tab:app-flow-released}
\end{table}

However, neither curvature nor action disagreement predicts benefit from additional steps across models. XR-0 and DiT have nearly identical metrics, yet XR-0 shows step-equivalence while DiT gains from multi-step integration; FLOWER also benefits from additional steps despite having a small geometric excess of $0.24\%$. SmolVLA provides a further counterexample: it is $3–4\times$ more curved, while having higher one-step performance. 

These results reinforce the findings of Sec.~\ref{app:appendix-action-space}: more similar flow geometry and closer flow trajectory endpoints between one- and multi-step inference do not necessarily lead to more similar success rates. One-step and multi-step inference can produce different actions while achieving equivalent task performance, like the XR-0 model. Thus, agreement in action space does not necessarily imply agreement in task success, or vice versa.

\section{Statistical tools}
\label{app:appendix-statistics}
\subsection{Equivalence Testing}

To claim a positive equivalence we test our hypothesis of step-invariance through the two one-sided $t$-tests (TOST) procedure~\cite{schuirmann1987comparison} (see Tab.~\ref{tab:tost_all_sim}). We conduct this test aggregating results across tasks, while for a per-task breakdown we refer to Appendix \ref{app:per-task-test}. For TOST we set a confidence level $\alpha=0.05$ and a margin of practical equivalence $\delta$. To claim equivalence, TOST needs to reject two composite null hypotheses, hence the CIs are computed at a confidence of $90\%$, i.e., $100(1-2\alpha)\%$. The procedure is identical for every metric: we estimate the gap $\Delta$ between the 1-step and reference-step settings, compute its $90\%$ CI, $[\Delta_L,\Delta_U]$, and apply the decision rule $\Pi$ of Eq.~\eqref{eq:tost-eq}. The only metric-dependent component is how the CI for $\Delta$ is computed, which differs between binary success outcomes and the CALVIN mean chain length.
\paragraph{Binary success outcomes.} For all success-rate metrics, including the CALVIN 5-task chain success (SR@5), we use a margin $\delta=3\,$pp (percentage points). Since per-rollout outcomes are binary, we treat the 1-step and reference-step methods as paired Bernoulli outcomes over matched rollouts. Using the paired contingency table (Tab.~\ref{tab:contingency}), we estimate the success rates $p_\textrm{ref},  p_\textrm{1}$ and the gap $\Delta$ as
\begin{equation}
    p_\textrm{1}=\frac{a+b}{N}\ , \quad p_\textrm{ref}=\frac{a+c}{N}\ , \quad \Delta=p_\textrm{ref}-p_\textrm{1}=\frac{c-b}{N}
\end{equation}
where we note that the gap only depends on the discordant pairs.
\begin{table}[ht]
\centering
    \caption{Contingency table. Each of the $N$ matched rollouts is classified by the joint outcome of the two inference settings: $a$ when both methods succeed, $b$ when 1-step succeeds but reference-step fails, $c$ when 1-step fails but reference-step succeeds, and $d$ when both fail.}
    \begin{tabular}{lcc}
    \toprule
     & reference-step Success& reference-step Failure  \\
    \midrule
    1-step Success & $a$ & $b$  \\
    1-step Failure  & $c$ & $d$  \\
    \midrule
    & \multicolumn{2}{c}{$N=a+b+c+d$}  \\
    \bottomrule
    \end{tabular}
\label{tab:contingency}
\end{table}

Since the two proportions are not independent due to matched rollouts, to compute the CI of the difference between the two means we adopt Newcombe's score-interval method for the difference between two paired proportions (method~10 of~\cite{newcombe1998improved}). This method computes the CI for $\Delta$ based on the Wilson score intervals~\cite{wilson1927probable} for $p_\textrm{1}$ and $p_\textrm{ref}$.
Let $[l_1,u_1]$ and $[l_\textrm{ref},u_\textrm{ref}]$ denote the Wilson score intervals at confidence level $1-2\alpha$ for $p_\textrm{1}$ and $p_\textrm{ref}$, respectively. The corresponding $100(1-2\alpha)\%$ CI $[\Delta_L,\Delta_U]$ for $\Delta$ is
\begin{equation}
    \begin{array}{cc}
\Delta_L &= \Delta - \sqrt{(p_\textrm{ref}-l_\textrm{ref})^2
   - 2\,\hat\phi\,(p_\textrm{ref}-l_\textrm{ref})(u_1-p_\textrm{1})
   + (u_1-p_\textrm{1})^2}\ ,\\ \\ 
\Delta_U &= \Delta + \sqrt{(u_\textrm{ref}-p_\textrm{ref})^2
   - 2\,\hat\phi\,(u_\textrm{ref}-p_\textrm{ref})(p_\textrm{1}-l_1)
   + (p_\textrm{1}-l_1)^2}\ ,
    \end{array}
\end{equation}

where $\hat\phi$ estimates the within-pair correlation. Writing ${\psi = ad-bc}$ and ${D=(a+b)(c+d)(a+c)(b+d)}$, $\hat\phi$ is defined as
\begin{equation}
\hat\phi =
\begin{cases}
(\psi - N/2)/\sqrt{D}\ , & \psi > N/2,\\[2pt]
0\ , & 0 \le \psi \le N/2,\\[2pt]
\psi/\sqrt{D}\ , & \psi < 0.
\end{cases}
\qquad (\hat\phi = 0 \text{ if } D=0) \ .
\end{equation}
\paragraph{CALVIN mean chain length.} For CALVIN we additionally test the mean chain length, i.e., the number of completed subtasks $s\in\{0,\dots,5\}$, which is no longer a Bernoulli outcome. For this test, we follow the procedure described in~\cite{lakens2017equivalence} to conduct an equivalence test for the differences between dependent means, and describe it using CIs instead of $p$ values. Let ${s^{\textrm{ref}}_i, s^{\textrm{1}}_i \in \{0,\dots,5\}}$ be the completion scores of the two settings on matched rollout $i$, and ${d_i = s^{\textrm{ref}}_i - s^{\textrm{1}}_i}$ their per-rollout difference. The gap is now the mean difference $\Delta=\bar d$ and, since both settings are evaluated on the same rollouts, we compute its CI as
\begin{equation}
\begin{array}{cc}
     \Delta_L &=  \Delta - t_{1-\alpha,N-1}\frac{s_d}{\sqrt{N}},\ \\
     \Delta_U &= \Delta + t_{1-\alpha,N-1}\frac{s_d}{\sqrt{N}},\
\end{array}
\end{equation}
where $t_{1-\alpha,N-1}$ is the upper-$\alpha$ quantile of the Student's $t$-distribution with $N-1$ degrees of freedom, and $s_d$ is the standard deviation $s_d=\sqrt{\frac{1}{N-1}\sum_{i=1}^{N}(d_i-\bar d)^2}$. To keep the margin comparable to the $3\,$pp used for the success rates, we set $\delta=0.15$, i.e. the same $3\%$, expressed on the $0$–$5$ scale.

Given $[\Delta_L,\Delta_U]$, in either case TOST declares the two methods equivalent within the pre-registered margin $\delta$ iff the CI for $\Delta$ is strictly contained in $[-\delta,\delta]$, we declare a difference if it lies completely outside of it, and otherwise the test is inconclusive. Formally, the TOST outcome $\Pi(\cdot)$ is defined as
\begin{equation}
    \Pi(\Delta_L,\Delta_U,\delta) =
    \begin{cases}
        \text{\textcolor{tostEqFg}{Equivalent}}
        & \text{if } [\Delta_L,\Delta_U] \subset (-\delta,\delta), \\[4pt]
        \textrm{\textcolor{tostDiffFg}{Different}}
        & \text{if } \Delta_L > \delta \ \lor\ \Delta_U < -\delta, \\[4pt]
        \text{\textcolor{tostIncFg}{Inconclusive}}
        & \text{otherwise.}
    \end{cases}
    \label{eq:tost-eq}
\end{equation}

\subsection{McNemar's test}
\label{app:mcnemar}

Since our real world evaluation setup is paired, i.e., the 1-step and reference-step approaches are tested under the same initial conditions and seeded policy noise, we can treat the binary outcome of each trial (success/failure) as paired observations and employ the two-sided McNemar test~\cite{mcnemar1947note} to assess if the difference in performance is statistically significant.

Given the contingency table Tab.~\ref{tab:contingency}, the null hypothesis $H_0$ is that the discordance is equally likely, i.e., $b=c$. As in standard practice, $p$-value computation changes depending on the sample size of discordant pairs: if there are no discordant pairs, $p=1.0$; for $b+c<25$ we use the exact binomial test
\begin{equation}
  p=
  2\sum_{i=0}^{\min(b,c)}
  \binom{b+c}{i}\left(\frac12\right)^{b+c} \quad ;
\end{equation}
when $b+c \geq 25$, we use chi-square approximation with Yates' correction for continuity
\begin{equation}
  \chi^2=\frac{(|b-c|-1)^2}{b+c},
  \qquad
  p=\Pr(\chi_1^2 \ge \chi^2) = 1-F_{\chi_1^2}(\chi^2)
\end{equation}
where $F_{\chi_1^2}(\chi^2)$ is the cumulative distribution function of the chi-squared distribution with 1 degree of freedom.

\textbf{Multiple-testing adjustment}. For aggregate results, we compute a single $p$-value as described above for each model and compare it directly to the significance level $\alpha=0.05$. For single task decision, to control the family-wise error rate at $\alpha$, we apply the Holm correction: given the raw $p$-value for each task computed as above, we first sort them as ($p_{(1)}\le\cdots\le p_{(4)}$), then compute the adjusted $p$-values as
\begin{equation}
  \begin{aligned}
  q_{(1)} &= \min(1,4p_{(1)}),\\
  q_{(2)} &= \min(1,\max(q_{(1)},3p_{(2)}))\, ,\\
  q_{(3)} &= \min(1,\max(q_{(2)},2p_{(3)}))\, ,\\
  q_{(4)} &= \min(1,\max(q_{(3)},p_{(4)})),\
  \end{aligned}
\end{equation}
and per-task decisions use $q_{(i)}<\alpha$.

\section{Different planning horizons}
\label{app:planning-horizons}
All the results reported in the main paper use the replanning horizon $H$, i.e., the number of actions in the current action chunk to execute before rolling out actions from the next chunk, chosen by the model authors for that task. Here, we test whether using a different $H$ would change our step equivalence claim (Tab.~\ref{tab:tost_horizon_all_sim}).

X-VLA, $\pi_{0.5}$ and XR-0 generally maintain equivalence at $H \geq 5$; LAP at the default of $H=5$ cannot claim equivalence on LIBERO-10, while at $H=10$ performance improves and the equivalence is established for every suite; for XR-0, on CALVIN we see less robustness to the choice of a smaller $H$ than the default value.
At $H=1$ multi-step generally outperforms one-step, with gains concentrated on long horizon tasks. However, at $H=1$, for all models except FLOWER, the multi-step policy still underperforms or matches the 1-step policy run at any longer horizon, so the choice of replanning at every timestep is in fact a net loss.
\begin{table}[t]
\centering
\caption{Replanning horizon $H$ analysis on LIBERO and CALVIN ABC$\rightarrow$D. Success rates and average chain length are reported along with TOST equivalence colouring (see Sec.~\ref{app:appendix-statistics}). SGO pools Spatial, Goal and Object suites; LIBERO-10 is its own suite. $\dagger$ marks that model's deployment default for that benchmark. LIBERO and CALVIN defaults differ for X-VLA (30 vs.\ 20), so those rows are benchmark-specific.}%
\label{tab:tost_horizon_all_sim}
\resizebox{\linewidth}{!}{%
\begin{tabular}{l*{9}{c}}
\toprule
\multirow{2}{*}{Model}
& \multirow{2}{*}{$h$}
& \multicolumn{2}{c}{LIBERO SGO}
& \multicolumn{2}{c}{LIBERO-10}
& \multicolumn{2}{c}{CALVIN SR@5}
& \multicolumn{2}{c}{CALVIN Avg. Len.} \\
\cmidrule(lr){3-4} \cmidrule(lr){5-6} \cmidrule(lr){7-8} \cmidrule(lr){9-10}
& & Reference & 1 Step & Reference & 1 Step & Reference & 1 Step & Reference & 1 Step \\
\midrule

\multirow{4}{*}{X-VLA} & 1 & \cellcolor{tostEqBg}61.6\% & \cellcolor{tostEqBg}\textbf{62.9\%} & \cellcolor{tostIncBg}10.6\% & \cellcolor{tostIncBg}\textbf{13.6\%} & \cellcolor{tostEqBg}\textbf{0.0\%} & \cellcolor{tostEqBg}0.0\% & \cellcolor{tostEqBg}\textbf{0.31} & \cellcolor{tostEqBg}0.30 \\
 & 10 & \cellcolor{tostEqBg}96.7\% & \cellcolor{tostEqBg}\textbf{97.2\%} & \cellcolor{tostEqBg}\textbf{94.4\%} & \cellcolor{tostEqBg}94.4\% & \cellcolor{tostIncBg}55.2\% & \cellcolor{tostIncBg}\textbf{56.4\%} & \cellcolor{tostEqBg}\textbf{3.64} & \cellcolor{tostEqBg}3.64 \\
 & 20\textsuperscript{$\dagger$} & -- & -- & -- & -- & \cellcolor{tostIncBg}\textbf{71.0\%} & \cellcolor{tostIncBg}70.0\% & \cellcolor{tostEqBg}\textbf{4.18} & \cellcolor{tostEqBg}4.14 \\
 & 30\textsuperscript{$\dagger$} & \cellcolor{tostEqBg}\textbf{97.6\%} & \cellcolor{tostEqBg}97.3\% & \cellcolor{tostEqBg}\textbf{96.2\%} & \cellcolor{tostEqBg}96.0\% & -- & -- & -- & -- \\
\midrule
\multirow{3}{*}{FLOWER} & 1 & \cellcolor{tostEqBg}\textbf{97.1\%} & \cellcolor{tostEqBg}96.7\% & \cellcolor{tostIncBg}\textbf{90.0\%} & \cellcolor{tostIncBg}85.0\% & \cellcolor{tostDiffBg}\textbf{46.7\%} & \cellcolor{tostDiffBg}4.8\% & \cellcolor{tostDiffBg}\textbf{3.48} & \cellcolor{tostDiffBg}1.44 \\
 & 5 & \cellcolor{tostEqBg}\textbf{97.4\%} & \cellcolor{tostEqBg}96.4\% & \cellcolor{tostIncBg}\textbf{90.0\%} & \cellcolor{tostIncBg}85.0\% & \cellcolor{tostDiffBg}\textbf{75.1\%} & \cellcolor{tostDiffBg}68.4\% & \cellcolor{tostDiffBg}\textbf{4.45} & \cellcolor{tostDiffBg}4.22 \\
 & 10\textsuperscript{$\dagger$} & \cellcolor{tostEqBg}\textbf{97.3\%} & \cellcolor{tostEqBg}96.5\% & \cellcolor{tostDiffBg}\textbf{89.6\%} & \cellcolor{tostDiffBg}83.6\% & \cellcolor{tostIncBg}\textbf{79.5\%} & \cellcolor{tostIncBg}77.0\% & \cellcolor{tostIncBg}\textbf{4.55} & \cellcolor{tostIncBg}4.43 \\
\midrule
\multirow{3}{*}{$\pi_{0.5}$} & 1 & \cellcolor{tostDiffBg}\textbf{94.2\%} & \cellcolor{tostDiffBg}88.9\% & \cellcolor{tostDiffBg}\textbf{85.6\%} & \cellcolor{tostDiffBg}64.2\% & -- & -- & -- & -- \\
 & 5\textsuperscript{$\dagger$} & \cellcolor{tostEqBg}98.5\% & \cellcolor{tostEqBg}\textbf{98.9\%} & \cellcolor{tostEqBg}\textbf{94.4\%} & \cellcolor{tostEqBg}94.4\% & -- & -- & -- & -- \\
 & 10 & \cellcolor{tostEqBg}\textbf{98.5\%} & \cellcolor{tostEqBg}98.4\% & \cellcolor{tostIncBg}\textbf{93.2\%} & \cellcolor{tostIncBg}92.2\% & -- & -- & -- & -- \\
\midrule
\multirow{3}{*}{LAP} & 1 & \cellcolor{tostIncBg}\textbf{95.1\%} & \cellcolor{tostIncBg}91.7\% & \cellcolor{tostIncBg}\textbf{82.4\%} & \cellcolor{tostIncBg}78.2\% & -- & -- & -- & -- \\
 & 5\textsuperscript{$\dagger$} & \cellcolor{tostEqBg}97.7\% & \cellcolor{tostEqBg}\textbf{98.8\%} & \cellcolor{tostIncBg}90.2\% & \cellcolor{tostIncBg}\textbf{92.0\%} & -- & -- & -- & -- \\
 & 10 & \cellcolor{tostEqBg}98.1\% & \cellcolor{tostEqBg}\textbf{98.2\%} & \cellcolor{tostEqBg}\textbf{93.4\%} & \cellcolor{tostEqBg}93.0\% & -- & -- & -- & -- \\
\midrule
\multirow{3}{*}{XR-0} & 1 & \cellcolor{tostIncBg}\textbf{91.7\%} & \cellcolor{tostIncBg}88.9\% & \cellcolor{tostDiffBg}\textbf{78.0\%} & \cellcolor{tostDiffBg}65.0\% & \cellcolor{tostDiffBg}\textbf{81.3\%} & \cellcolor{tostDiffBg}74.5\% & \cellcolor{tostDiffBg}\textbf{4.58} & \cellcolor{tostDiffBg}4.29 \\
 & 5 & \cellcolor{tostEqBg}\textbf{98.7\%} & \cellcolor{tostEqBg}98.1\% & \cellcolor{tostEqBg}92.8\% & \cellcolor{tostEqBg}\textbf{93.4\%} & \cellcolor{tostIncBg}\textbf{88.7\%} & \cellcolor{tostIncBg}86.7\% & \cellcolor{tostEqBg}\textbf{4.77} & \cellcolor{tostEqBg}4.70 \\
 & 10\textsuperscript{$\dagger$} & \cellcolor{tostEqBg}\textbf{97.3\%} & \cellcolor{tostEqBg}97.0\% & \cellcolor{tostEqBg}96.8\% & \cellcolor{tostEqBg}\textbf{97.4\%} & \cellcolor{tostEqBg}87.2\% & \cellcolor{tostEqBg}\textbf{88.1\%} & \cellcolor{tostEqBg}4.73 & \cellcolor{tostEqBg}\textbf{4.73} \\

\bottomrule
\end{tabular}%
}
\end{table}

\section{Experimental details}
\subsection{Real world experiment details}
\label{app:real-world-finetuning}
\textbf{Dataset}:
Our in-house dataset is composed of \num{68872} frames spanning $410$ episodes and $9$ tasks, collected at a frequency of \SI{12.5}{\Hz} by a single human operator. 
Tab.~\ref{tab:dataset-info} lists the prompt and duration associated with each task.
Every task is a tabletop task, and together they cover a range of contact-rich interactions: opening and closing hinge joints (the oven) and slide joints (the drawer), precision manipulation in the \textit{Open oven} task, as well as simpler pick-and-place tasks.
The dataset is deliberately designed to require language grounding, i.e., given the same initial state multiple tasks can be carried out depending on the input prompt. Although the dataset consists of 9 tasks, in this work we evaluate only on a subset of 4 tasks. The training dataset is the same as the one used in~\cite{zhang2026augmented}.

\textbf{Model Training}:
Each training sample combines the robot proprioceptive information with three camera images: a front view, a wrist view, and a left egocentric view.
The proprioceptive state of the robot is $\left(\bm{x}, \bm{r}, g\right)$, where $\bm{x} \in \mathbb{R}^3$ is the end effector position, $\bm{r} \in \mathbb{R}^3$ is the end effector orientation in axis-angle representation, and $g \in \left[ 0, 1\right]$ is the gripper state.
The action space is $\left(\Delta\bm{x}, \Delta\bm{r}, g_{cmd}\right)$, where $\Delta\bm{x}, \Delta\bm{r}, g_{cmd}$ are deltas to the end effector pose and the gripper command, respectively.
The batch size and the number of gradient steps used for fine-tuning are reported in Tab.~\ref{tab:deploy_train_hparams}.
To avoid introducing confounders, we reuse the hyperparameters reported by the authors of each method for the post-training experiments.

\textbf{Policy Deployment}:
In Tab~\ref{tab:deploy_train_hparams} we report the hyperparameters used for our real world experiments, all of which run at a control frequency of \SI{12.5}{\Hz}. During deployment we observed that a replanning horizon of $10$ degraded performance for X-VLA, FLOWER, and DiT, so we increased this horizon on a per-model basis until we obtained smoother trajectories. The reported parameters were kept the same for the one-step and reference-step variants.
Across all the experiments, a trial was counted as a failure when either a timeout of $400$ steps was reached, when the robot got into unrecoverable error cases, or when poor language following caused the policy to carry out a task other than the one specified in the prompt. The evaluation budget is reported in Tab.~\ref{tab:benchmark_overview}.

\begin{table}[t]
\centering
\caption{Median episode length per task in the Real World dataset (410 total episodes). Lengths are reported in frames, with the corresponding duration in seconds in parentheses.}
\label{tab:dataset-info}
\begin{tabular}{lr}
\toprule
Task prompt & Median length in frames (seconds) \\
\midrule
Close the drawer                       & 106   (8.83) \\
Close the oven                         & 113.5 (9.46) \\
Open the drawer                        & 177.5 (14.79) \\
Open the oven                          & 119   (9.92) \\
Put the green plate in the oven        & 197   (16.42) \\
Put the mug in the white basket        & 195   (16.25) \\
Put the spoon in the drawer            & 170   (14.17) \\
Put the spoon in the yellow plate      & 257   (21.42) \\
Take the mug and give it to the user   & 187   (15.58) \\
\bottomrule
\end{tabular}
\end{table}

\begin{table*}[t]
\centering
\caption{
Training and deployment hyperparameters.
Budget denotes the total number of fine-tuning gradient steps, while
checkpoint denotes the checkpoint used for evaluation.
We also report the action chunk size and the replanning horizon.
}
\label{tab:deploy_train_hparams}
\small
\setlength{\tabcolsep}{5pt}
\begin{tabular}{@{}llrrrrr@{}}
\toprule
Model & Benchmark & Budget & Checkpoint & Batch & Chunk & Replan \\
\midrule

\multirow{3}{*}{DiT}
 & Real World    & \num{30000} & \num{30000} & 256 & 25 & 15 \\
 & LIBERO        & \num{30000} & \num{30000} & 256 & 25 & 10 \\
 & D3IL Avoiding & \num{50000} & \num{50000} & 256 & 10 & 10 \\
\midrule

\multirow{2}{*}{X-VLA}
 & Real World    & \num{50000} & \num{50000} & 128 & 25 & 25 \\
 & D3IL Avoiding & \num{10000} & \num{10000} & 128 & 10 & 10 \\
\midrule

\multirow{2}{*}{FLOWER}$^\dagger$
 & Real World    & \num{40000} & \num{23000} & 32 & 25 & 15 \\
 & D3IL Avoiding & \num{10000} & \num{5000}            & 32 & 10 & 10 \\
\midrule

\multirow{2}{*}{$\pi_{0.5}$}
 & Real World    & \num{30000} & \num{30000} & 128 & 25 & 10 \\
 & D3IL Avoiding & \num{20000} & \num{5000}  & 256 & 10 & 10 \\
\midrule

\multirow{2}{*}{LAP}
 & Real World    & \num{40000} & \num{40000} & 256 & 25 & 10 \\
 & D3IL Avoiding & \num{10000} & \num{5000}  & 256 & 10 & 10 \\
\midrule

\multirow{2}{*}{XR-0}
 & Real World    & \num{30000} & \num{30000} & 128 & 30 & 10 \\
 & D3IL Avoiding & \num{10000} & \num{10000} & 256 & 10 & 10 \\

\bottomrule
\end{tabular}
\begin{minipage}{0.96\linewidth}
\footnotesize
$^\dagger$ For FLOWER we use early stopping as in the original work and hence evaluate earlier checkpoints based on the action loss.
\end{minipage}
\end{table*}

\subsection{D3IL setup}
\label{app:appendix-d3il}
We consider only the Avoiding task in D3IL benchmark: the training dataset from the authors is state based and is made of 24 successful trajectories with 4 demonstrations each (Tab.~\ref{tab:benchmark_overview}). We replay those demonstrations and collect images to build the training dataset to train our models on. As reported in Tab.~\ref{tab:deploy_train_hparams}, we do not always take the last checkpoint when training models on D3IL. In Tab.~\ref{tab:d3il_checkpoint_sr_entropy}, we report the success rate and the entropy across models and checkpoints. All conditions use 200 rollouts, an action chunk and replanning horizon of 10, and a maximum episode length of 150 steps. Entropy is computed over successful trajectory modes and normalised $\in [0, 1]$ by dividing by $\log(24)$.

\begin{table}[t]
\centering
\caption{D3IL Avoiding success rate and normalised entropy across evaluated
training checkpoints. The checkpoints that were used for the main results are highlighted in grey.}
\label{tab:d3il_checkpoint_sr_entropy}
\begin{tabular}{llcccc}
\toprule
\multirow{2}{*}{Model}
& \multirow{2}{*}{Gradient Step}
& \multicolumn{2}{c}{Success Rate (\%) $\uparrow$}
& \multicolumn{2}{c}{Normalised Entropy $\uparrow$} \\
\cmidrule(lr){3-4}
\cmidrule(lr){5-6}
& & 1 Step & Reference & 1 Step & Reference \\
\midrule

\multirow{5}{*}{DiT}
& \num{10000} & 4.0 & \textbf{76.5} & 0.000 & \textbf{0.045} \\
& \num{20000} & 19.5 & \textbf{79.5} & 0.000 & 0.000 \\
& \num{30000} & 78.5 & \textbf{98.0} & \textbf{0.218} & 0.037 \\
& \num{40000} & 86.5 & \textbf{99.5} & \textbf{0.101} & 0.000 \\
& \cellcolor{gray!15}\num{50000}
& \cellcolor{gray!15}\textbf{100.0}
& \cellcolor{gray!15}70.0
& \cellcolor{gray!15}0.000
& \cellcolor{gray!15}\textbf{0.148} \\
\midrule

\multirow{2}{*}{X-VLA}
& \num{5000} & \textbf{62.5} & 54.0 & \textbf{0.376} & 0.363 \\
& \cellcolor{gray!15}\num{10000}
& \cellcolor{gray!15}79.0
& \cellcolor{gray!15}\textbf{89.5}
& \cellcolor{gray!15}\textbf{0.051}
& \cellcolor{gray!15}0.019 \\
\midrule

FLOWER
& \cellcolor{gray!15}\num{5000}
& \cellcolor{gray!15}\textbf{99.5}
& \cellcolor{gray!15}12.5
& \cellcolor{gray!15}0.000
& \cellcolor{gray!15}\textbf{0.167} \\
\midrule

\multirow{4}{*}{$\pi_{0.5}$}
& \cellcolor{gray!15}\num{5000}
& \cellcolor{gray!15}87.0
& \cellcolor{gray!15}\textbf{88.5}
& \cellcolor{gray!15}0.286
& \cellcolor{gray!15}\textbf{0.737} \\
& \num{10000} & \textbf{93.0} & 87.0 & 0.191 & \textbf{0.857} \\
& \num{15000} & 92.5 & 92.5 & 0.154 & \textbf{0.538} \\
& \num{20000} & 73.5 & \textbf{89.0} & \textbf{0.562} & 0.515 \\
\midrule

\multirow{2}{*}{LAP}
& \cellcolor{gray!15}\num{5000}
& \cellcolor{gray!15}\textbf{100.0}
& \cellcolor{gray!15}95.0
& \cellcolor{gray!15}0.000
& \cellcolor{gray!15}\textbf{0.140} \\
& \num{10000} & 71.5 & \textbf{86.0} & 0.356 & \textbf{0.568} \\
\midrule

\multirow{2}{*}{XR-0}
& \num{5000} & 48.0 & \textbf{66.0} & 0.361 & \textbf{0.537} \\
& \cellcolor{gray!15}\num{10000}
& \cellcolor{gray!15}\textbf{91.0}
& \cellcolor{gray!15}83.5
& \cellcolor{gray!15}0.373
& \cellcolor{gray!15}\textbf{0.499} \\
\bottomrule
\end{tabular}
\end{table}

\begin{table}[t]
\centering
\caption{Benchmark Overview. Eval. Budget refers to the number of per-task rollouts.}
\label{tab:benchmark_overview}
\begin{threeparttable}
\resizebox{\linewidth}{!}{%
\begin{tabular}{lcccc}
\midrule
Benchmark & Splits & Data Samples & Tasks & Eval. Budget \\
\midrule
LIBERO~\cite{liu2023libero} & Spatial, Object, Goal (SGO); LIBERO 10 & $273{,}465$ & $40$ & $50$ \\
LIBERO+~\cite{fei2025liberop} & Spatial, Object, Goal (SGO); LIBERO 10 & $2{,}238{,}036$ & $40$ & $\sim250$ \\
CALVIN~\cite{mees2022calvin} & ABC$\to$D~\cite{mees2022calvin} & $1{,}071{,}743$ & $389$ & $1000$\tnote{1} \\
RoboTwin~\cite{chen2025robotwin} & Clean, Randomised & $248{,}105$ & $50$ & $100$ \\
Real world~\cite{zhang2026augmented} & - & $68{,}872$ & $9$ & $100$ \\
D3IL Avoiding~\cite{jia2024towards} & - & $96$ & $1$ & $200$ \\
\bottomrule
\end{tabular}%
}
\begin{tablenotes}
  \item[1] CALVIN does not evaluate per task, but instead on 1000 chains of 5 randomly chosen tasks.
\end{tablenotes}
\end{threeparttable}
\end{table}

\section{Per-task testing}
\label{app:per-task-test}

Following the approach in~\cite{barreiros2026careful}, we use violin plots to visualise the uncertainty in the success rate estimation. 
The violin plots represent a Bayesian posterior probability under a uniform Beta prior given the observed success rate and total number of runs. We model the binary outcome ${y_i \sim \textrm{Bernoulli}(\theta)}$ and $\theta \sim \textrm{Beta}(1,1)$. Given N rollouts with binary outcomes $Y=\{y_1, \hdots ,y_{N}\}$, assuming we recorded $s$ successes in $N$ trials, the posterior probability is
\begin{equation}
    p(\theta|Y) \propto \textrm{Beta}(s+1, N-s+1),
    \label{eq:beta-posterior}
\end{equation}
and the violin plots are made with points sampled from Eq.~\eqref{eq:beta-posterior}.


The comparison is carried out using the STEP framework~\cite{snyder2025your}, with the Bonferroni correction over the per-task family of hypotheses controlling the family-wise error rate at $\alpha_{\text{global}} = 0.05$, hence running task-level tests with $\alpha_\textrm{split}=\alpha_\textrm{global}/K$ where $K$ is the family size. We note that we use Bonferroni correction here instead of Holm correction as in Sec. \ref{app:mcnemar} since the STEP framework does not return a $p$-value like McNemar, but a decision at a given $\alpha$. We use $K=10$ for LIBERO and LIBERO+, $K=50$ for RoboTwin, $K=4$ for the Real World suite, while for CALVIN we do not apply Bonferroni correction.

In the violin plots, we use compact letter display to mark a statistically significant difference. Violins with different letters and marked with $\star \star$ were separable at the given level of confidence $\alpha_\textrm{split}$ after Bonferroni correction. Violins sharing the same letter but with a single $\star$, are separable at $\alpha_\textrm{global}=0.05$ but are not after Bonferroni correction. Violins that share the same letter and do not have a $\star$ are not separable.

\subsection{CALVIN}
\label{app:appendix-statistics-calvin}
Evaluation for the CALVIN datasets consists of \num{1000} chains of $5$ randomly chosen tasks.
We report in Tab.~\ref{tab:calvin_results_glyph} the success rate of 1 up to 5 consecutive tasks, and the average chain length.
We compare the full chain success rate SR@5 with STEP, without Bonferroni correction, and the two methods are not separable with this evaluation budget for any VLA.
For the Avg. Len. we run a Welch t-test and FLOWER shows a significant drop with $p=0.013$.
\begin{table}[htp]
\centering
\caption{CALVIN ABC-D success rates for 1-step and reference-step variants. Glyphs on SR@5 and Avg. Len. encode the 1-step vs reference-step comparison: $\textcolor{OliveGreen}{\blacktriangle}$ 1-step significantly better, $\textcolor{OrangeRed}{\blacktriangledown}$ 1-step significantly worse. Avg. Len. uses a Welch t-test at ${p < 0.05}$.}
\label{tab:calvin_results_glyph}
\resizebox{\linewidth}{!}{%
\begin{tabular}{lcccccccccccc}
\toprule
\multirow{2}{*}{Model} & \multicolumn{2}{c}{SR@1} & \multicolumn{2}{c}{SR@2} & \multicolumn{2}{c}{SR@3} & \multicolumn{2}{c}{SR@4} & \multicolumn{2}{c}{SR@5} & \multicolumn{2}{c}{Avg. Len.} \\
\cmidrule(lr){2-3} \cmidrule(lr){4-5} \cmidrule(lr){6-7} \cmidrule(lr){8-9} \cmidrule(lr){10-11} \cmidrule(lr){12-13}
 & Reference & 1-step & Reference & 1-step & Reference & 1-step & Reference & 1-step & Reference & 1-step & Reference & 1-step \\
\midrule
X-VLA & \textbf{95.8\%} & 95.1\% & \textbf{89.8\%} & 88.9\% & \textbf{83.8\%} & 83.1\% & \textbf{77.9\%} & 77.3\% & \textbf{71.0\%} & 70.0\% & \textbf{4.18} & 4.14 \\
FLOWER & \textbf{99.5\%} & 99.3\% & \textbf{96.1\%} & 95.2\% & \textbf{91.6\%} & 89.9\% & \textbf{85.8\%} & 83.2\% & \textbf{78.5\%} & 75.5\% & \textbf{4.51} & 4.43 {\scriptsize $\textcolor{OrangeRed}{\blacktriangledown}$} \\
XR-0 & 99.9\% & 99.9\% & \textbf{98.1\%} & 97.8\% & \textbf{95.6\%} & 95.4\% & 92.2\% & 92.2\% & 87.2\% & \textbf{88.1\%} & 4.73 & 4.73 \\
\bottomrule
\end{tabular}
}
\end{table}
\subsection{LIBERO and LIBERO+}
\label{app:appendix-statistics-libero}

\begin{table}[htp]
\centering
\caption{LIBERO and LIBERO+ success rates. SGO is an average over Spatial, Goal, and Object suites. 
}
\label{tab:libero_liberoplus_compact_glyph}
\resizebox{\linewidth}{!}{%
\begin{tabular}{lcccccccc}
\midrule
\multirow{2}{*}{Model} 
& \multicolumn{2}{c}{LIBERO SGO}
& \multicolumn{2}{c}{LIBERO-10}
& \multicolumn{2}{c}{LIBERO+ SGO}
& \multicolumn{2}{c}{LIBERO+ 10} \\
\cmidrule(lr){2-3}
\cmidrule(lr){4-5}
\cmidrule(lr){6-7}
\cmidrule(lr){8-9}
& Reference & 1 Step & Reference & 1 Step & Reference & 1 Step & Reference & 1 Step \\
\midrule
DiT 
& 89.9\% & \textbf{91.9\%} {\scriptsize $\textcolor{OliveGreen}{\blacktriangle}\textcolor{OrangeRed}{\blacktriangledown}$}
& \textbf{68.4\%} & 53.8\% {\scriptsize $\textcolor{OrangeRed}{\blacktriangledown}$}
& \textbf{30.8\%} & 28.6\%
& \textbf{22.0\%} & 14.5\% {\scriptsize $\textcolor{OrangeRed}{\blacktriangledown}$} \\

SmolVLA 
& 72.2\% & \textbf{82.0\%} {\scriptsize $\textcolor{OliveGreen}{\blacktriangle}\textcolor{OrangeRed}{\blacktriangledown}$}
& 34.6\% & \textbf{44.2\%} {\scriptsize $\textcolor{OliveGreen}{\blacktriangle}$}
& 4.8\% & \textbf{5.0\%} {\scriptsize $\textcolor{OrangeRed}{\blacktriangledown}$}
& \textbf{1.2\%} & 0.7\% \\

X-VLA 
& \textbf{97.6\%} & 97.3\%
& \textbf{96.2\%} & 96.0\%
& 70.3\% & \textbf{70.4\%}
& \textbf{67.4\%} & 66.9\% \\

FLOWER 
& \textbf{97.3\%} & 96.5\%
& \textbf{89.6\%} & 83.6\%
& \textbf{77.6\%} & 74.4\% {\scriptsize $\textcolor{OrangeRed}{\blacktriangledown}$}
& \textbf{72.0\%} & 59.3\% {\scriptsize $\textcolor{OrangeRed}{\blacktriangledown}$} \\

$\pi_{0.5}$ 
& 98.5\% & \textbf{98.9\%}
& 94.4\% & 94.4\%
& \textbf{86.3\%} & 86.2\%
& \textbf{77.7\%} & 76.3\% {\scriptsize $\textcolor{OrangeRed}{\blacktriangledown}$} \\

LAP 
& 97.7\% & \textbf{98.8\%}
& 90.2\% & \textbf{92.0\%}
& \textbf{77.6\%} & 76.9\%
& \textbf{66.4\%} & 65.7\% \\

XR-0 
& \textbf{97.3\%} & 97.0\%
& 96.8\% & \textbf{97.4\%}
& 75.9\% & \textbf{76.6\%}
& \textbf{70.4\%} & 69.7\% \\
\bottomrule
\end{tabular}
}
\end{table}

We consider four suites of the LIBERO and LIBERO+ benchmark: Spatial, Goal, Object (SGO) and LIBERO 10. Within each of these suites, there are 10 subtasks, and for our evaluation we perform 50 rollouts per task. We compare each model doing 1-step inference and multi-step inference with the number of steps as reported in the original paper.

\begin{figure}[h]
    \centering
    \includegraphics[width=\linewidth]{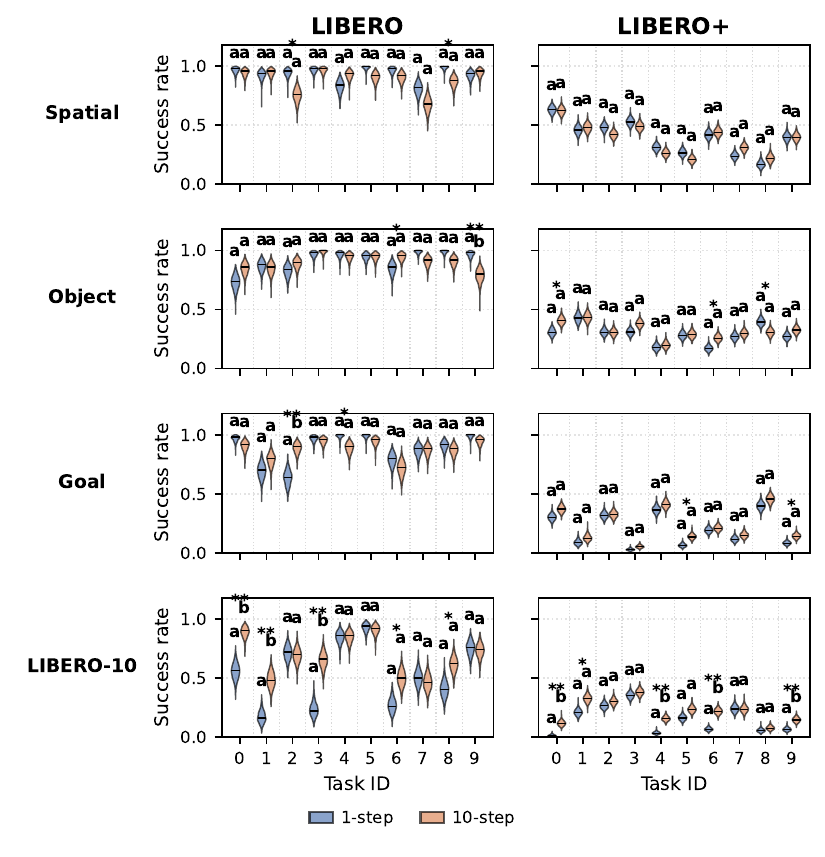}
    \caption{DiT per-task analysis on LIBERO. Violins sharing the same letter but with a single $\star$, are separable at $\alpha_\textrm{global}=0.05$ but are not after Bonferroni correction.}
    \label{fig:dit-libero}
\end{figure}

\textbf{DiT}: For DiT (see Fig.~\ref{fig:dit-libero}), in the SGO suites there is not a setting that is clearly better than the other, with each setting being better at 3 out of the 6 separable tasks before Bonferroni correction for LIBERO. For the long horizon LIBERO 10, the 10-step approach is the clear winner for this small scale model, especially in the LIBERO+ setting, and it aligns with our findings. 

\begin{figure}[h]
    \centering
    \includegraphics[width=\linewidth]{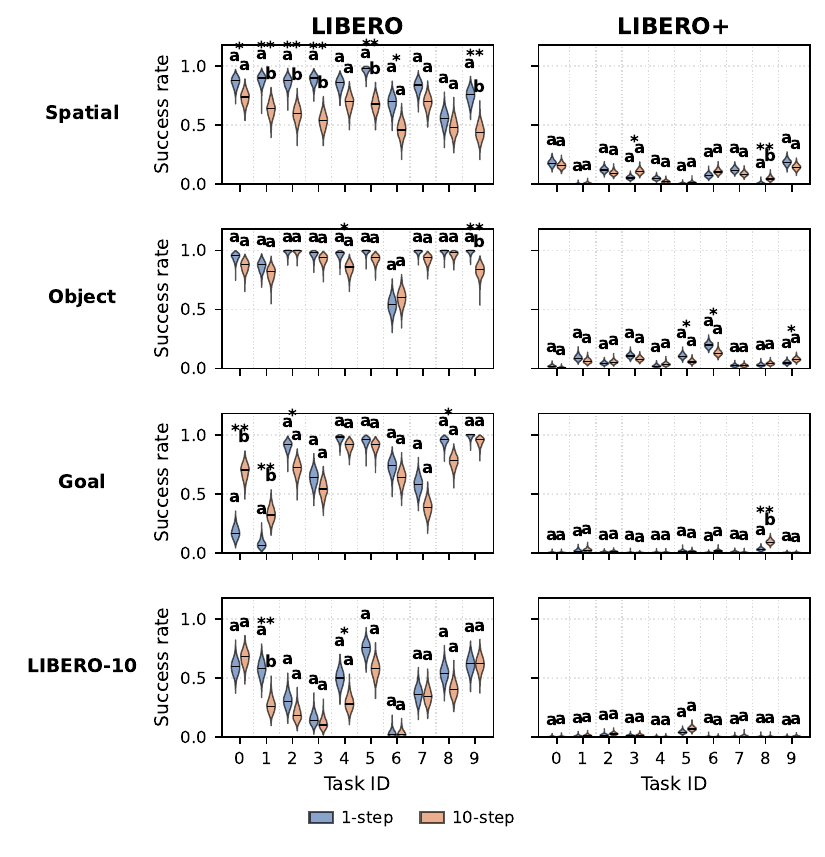}
    \caption{SmolVLA per-task analysis on LIBERO. Violins sharing the same letter but with a single $\star$, are separable at $\alpha_\textrm{global}=0.05$ but are not after Bonferroni correction.}
    \label{fig:smolvla-libero}
\end{figure}

\textbf{SmolVLA}: For SmolVLA (see Fig.~\ref{fig:smolvla-libero}), in LIBERO there are 7 tasks where the 1-step is significantly better than the 10-step approach, while the opposite happens only twice in the Goal suite.
The signal for this model is quite strong: since multi-step Euler integration better approximates the integration of the learned velocity field, a worse performance would indicate that the 1-step approach learns a field that poorly approximates the target.
Therefore, integrating it more faithfully hardly improves the performance, and can actually be detrimental.
For LIBERO+ the model performance is near zero so the information signal is low.

\begin{figure}[h]
    \centering
    \includegraphics[width=\linewidth]{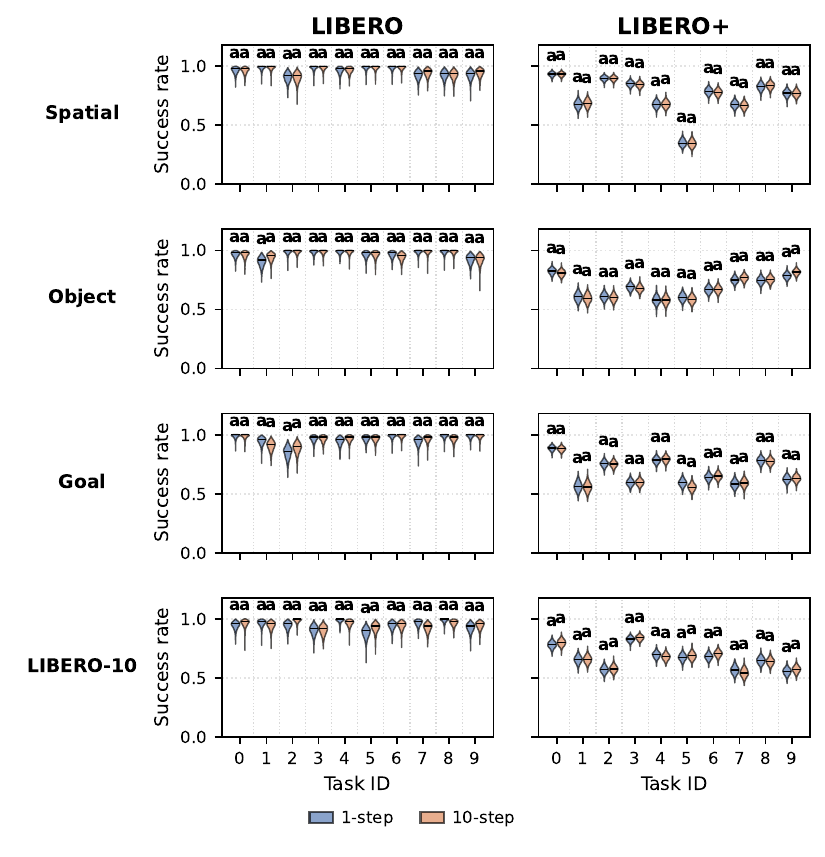}
    \caption{X-VLA per-task analysis on LIBERO. Violins sharing the same letter but with a single $\star$, are separable at $\alpha_\textrm{global}=0.05$ but are not after Bonferroni correction.}
    \label{fig:xvla-libero}
\end{figure}

\textbf{X-VLA, $\bm{\pi_{0.5}}$\,, XR-0}: For X-VLA and XR-0 (see Fig.~\ref{fig:xvla-libero} and Fig.~\ref{fig:XR-0-libero}), none of the tasks is separable even at $\alpha_\textrm{global}$. For $\pi_{0.5}$ (Fig.~\ref{fig:pi05-libero}), only one task in the long horizon and distribution shifted LIBERO+ 10 suite shows a separation.

\begin{figure}[h]
    \centering
    \includegraphics[width=\linewidth]{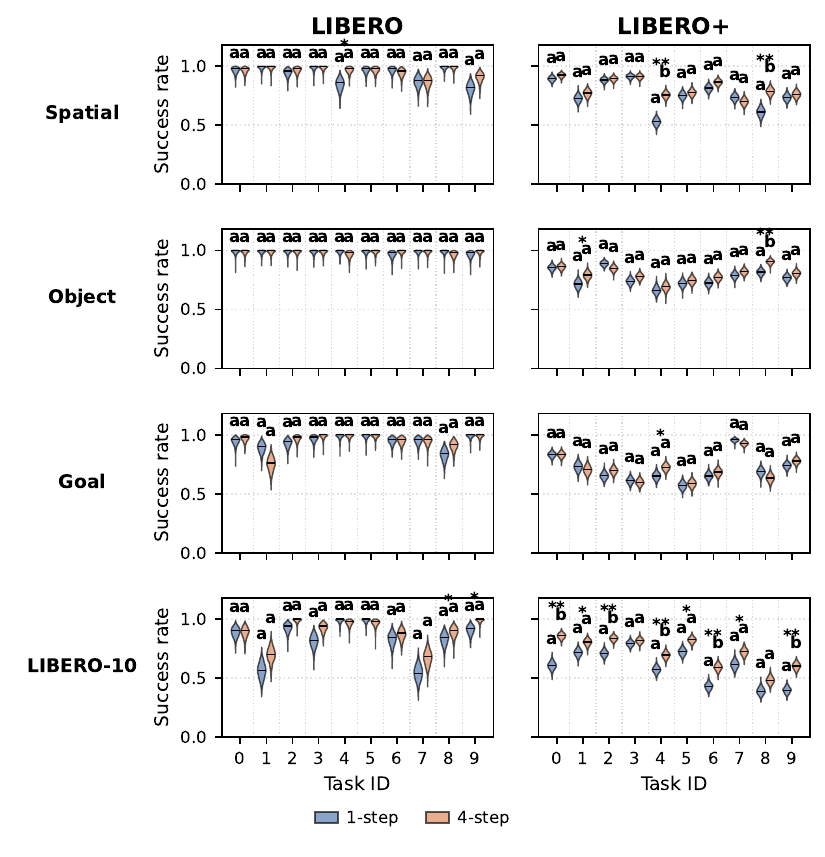}
    \caption{FLOWER per-task analysis on LIBERO. Violins sharing the same letter but with a single $\star$, are separable at $\alpha_\textrm{global}=0.05$ but are not after Bonferroni correction.}
    \label{fig:flower-libero}
\end{figure}

\begin{figure}[h]
    \centering
    \includegraphics[width=\linewidth]{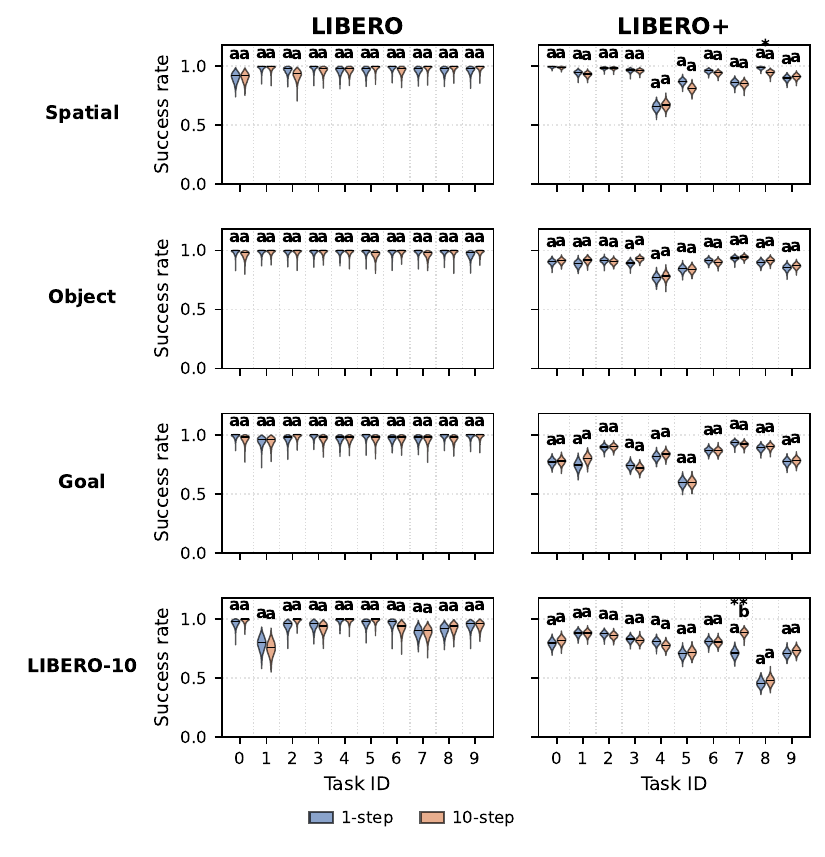}
    \caption{$\pi_{0.5}$ per-task analysis on LIBERO. Violins sharing the same letter but with a single $\star$, are separable at $\alpha_\textrm{global}=0.05$ but are not after Bonferroni correction.}
    \label{fig:pi05-libero}
\end{figure}

\textbf{FLOWER}: For FLOWER (see Fig.~\ref{fig:flower-libero}), no task survives Bonferroni correction in the LIBERO benchmark, but three of them are separable before correction in favour of the multi-step approach. This signal is stronger in the LIBERO+ benchmark, especially in the long horizon suite, corroborating our findings of long horizon and small scale being a setting where the two approaches most often differ.

\begin{figure}[h]
    \centering
    \includegraphics[width=\linewidth]{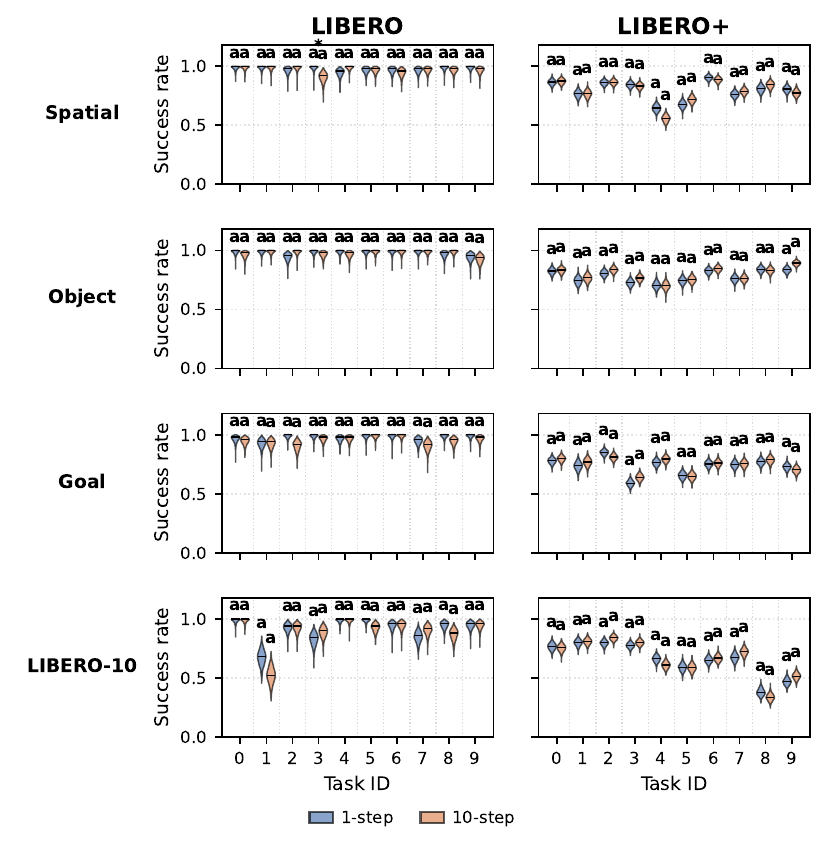}
    \caption{LAP per-task analysis on LIBERO. Violins sharing the same letter but with a single $\star$, are separable at $\alpha_\textrm{global}=0.05$ but are not after Bonferroni correction.}
    \label{fig:lap-libero}
\end{figure}

\begin{figure}[h]
    \centering
    \includegraphics[width=\linewidth]{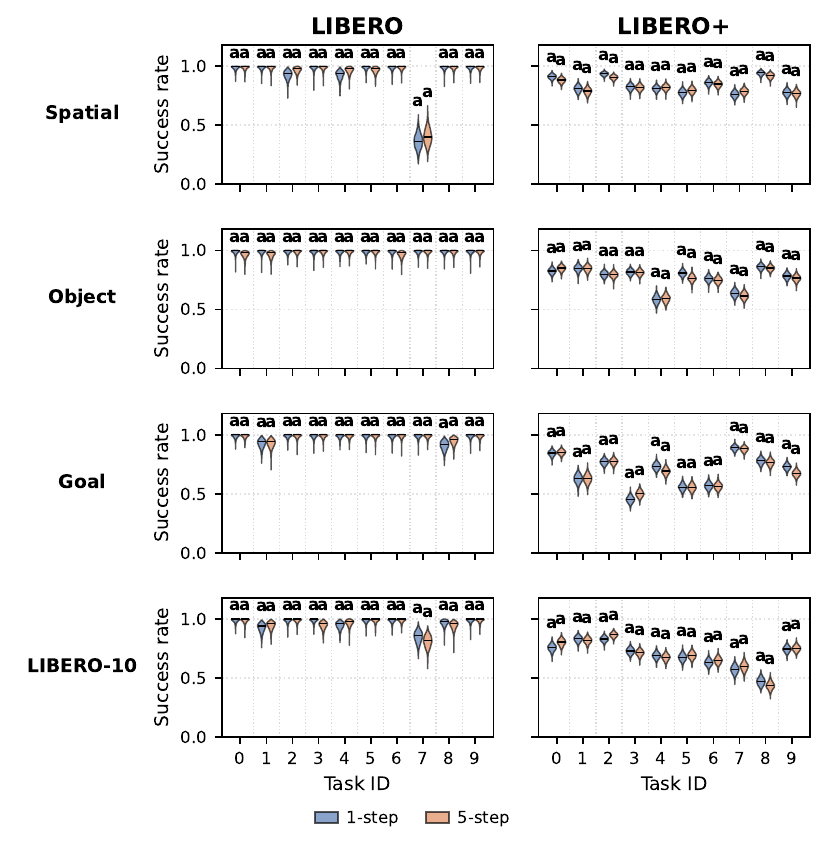}
    \caption{XR-0 per-task analysis on LIBERO. Violins sharing the same letter but with a single $\star$, are separable at $\alpha_\textrm{global}=0.05$ but are not after Bonferroni correction.}
    \label{fig:XR-0-libero}
\end{figure}

\textbf{LAP}: For LAP (see Fig.~\ref{fig:lap-libero}) only one task is separable at $\alpha_\textrm{global}$ in favour of the 1-step approach. The 1-step approach is also marginally better on the aggregate results.



\end{document}